\documentclass{article}
\usepackage[final]{neurips_2023}
\pdfoutput=1

\usepackage{microtype}
\usepackage{graphicx}
\usepackage{subfigure}
\usepackage{booktabs}
\usepackage{arydshln}
\usepackage{multirow}
\usepackage{multicol} 
\usepackage{makecell}
\usepackage{etoolbox}
\makeatletter
\usepackage{wrapfig}
\usepackage{caption}
\usepackage{amsmath,amssymb,amsfonts}
\usepackage{wrapfig}
\usepackage{microtype}
\usepackage{graphicx}
\usepackage{subfigure}
\usepackage{booktabs}
\usepackage{ulem}
\usepackage{hyperref}

\usepackage[ruled]{algorithm2e}
\SetAlgoCaptionSeparator{}

\usepackage[utf8]{inputenc} 
\usepackage[T1]{fontenc}    
\usepackage{hyperref}       
\usepackage{url}            
\usepackage{booktabs}       
\usepackage{amsfonts}       
\usepackage{nicefrac}       
\usepackage{microtype}      
\usepackage{xcolor}         

\usepackage[capitalize,noabbrev]{cleveref}

\usepackage[textsize=tiny]{todonotes}

\title{MathNAS: If Blocks Have a Role in Mathematical Architecture Design
}
\author{%
\textbf{Qinsi Wang}$^1$\thanks{Equal contribution.} \quad \quad \textbf{Jinghan Ke}$^{1*}$ \quad \quad \textbf{Zhi Liang}$^{2}$ \quad \quad \textbf{Sihai Zhang} $^{3,4}$\\
     ${}^{1}$ \text{University of Science and Technology of China}\\
     ${}^{2}$ \text{School of Life Sciences, University of Science and Technology of China}\\
     ${}^{3}$ \text{Key Laboratory of Wireless-Optical Communications, Chinese Academy of Sciences}\\
     ${}^{4}$ \text{School of Microelectronics, University of Science and Technology of China}\\
  \texttt{\{wqs,jinghan\}@mail.ustc.edu.cn},
  \texttt{\{liangzhi,shzhang\}@ustc.edu.cn}\\
}

\begin{document}

\maketitle

\begin{abstract}
Neural Architecture Search (NAS) has emerged as a favoured method for unearthing effective neural architectures. 
Recent development of large models has intensified the demand for faster search speeds and more accurate search results. 
However, designing large models by NAS is challenging due to the dramatical increase of search space and the associated huge performance evaluation cost. 
Consider a typical modular search space widely used in NAS, in which a neural architecture consists of $m$ block nodes and a block node has $n$ alternative blocks. 
Facing the space containing $n^m$ candidate networks, existing NAS methods attempt to find the best one by searching and evaluating candidate networks directly.
Different from the general strategy that takes architecture search as a whole problem, we propose a novel divide-and-conquer strategy by making use of the modular nature of the search space.
Here, we introduce MathNAS, a general NAS framework based on mathematical programming. 
In MathNAS, the performances of the $m*n$ possible building blocks in the search space are calculated first, and then the performance of a network is directly predicted based on the performances of its building blocks.
Although estimating block performances involves network training, just as what happens for network performance evaluation in existing NAS methods, predicting network performance is completely training-free and thus extremely fast. In contrast to the $n^m$ candidate networks to evaluate in existing NAS methods, which require training and a formidable computational burden, there are only $m*n$ possible blocks to handle in MathNAS.
Therefore, our approach effectively reduces the complexity of network performance evaluation. 
The superiority of MathNAS is validated on multiple large-scale CV and NLP benchmark datasets. 
Notably on ImageNet-1k, MathNAS achieves 82.5\% top-1 accuracy, 1.2\% and 0.96\% higher than Swin-T and LeViT-256, respectively. 
In addition, when deployed on mobile devices, MathNAS achieves real-time search and dynamic network switching within 1s (0.4s on TX2 GPU), surpassing baseline dynamic networks in on-device performance.
Our code is available at https://github.com/wangqinsi1/MathNAS.
\end{abstract}

\section{Introduction}
\label{introduction}
Neural Architecture Search (NAS) has notably excelled in designing efficient models for Computer Vision (CV) \cite{1,2,3,4} and Natural Language Processing (NLP) \cite{5,6,7} tasks. 
With the growing popularity of the Transformer architecture \cite{8,9}, designers are increasingly drawn to using NAS to develop powerful large-scale models. 
Many existing NAS studies focus on designing search spaces for large models and conducting searches within them \cite{10,11,12}.

However, designing large models by NAS is challenging due to the dramatical increase of search space and the associated huge performance evaluation cost \cite{10,13}.
Consider a widely used modular search space, in which a neural architecture is treated as a topological organization of $m$ different block nodes and each block node has $n$ different block implementations. 
Obviously, the number of possible networks or neural architectures, $n^m$, grows explosively with $n$ and $m$.
In addition, candidate networks of large models are larger and require more computation for performance evaluation.
Therefore, in order to conduct an effective architecture search, a proper search strategy and a suitable performance evaluation method are extremely important.

It is noteworthy that to improve search strategy, recent researches \cite{14,15} convert NAS to mathematical programming (MP) problems, which substantially decrease the search cost. MP-NAS provides a promising direction for rapidly designing large models. 
However, current MP-NAS methods exhibit certain architectural constraints. 
For example, DeepMAD \cite{14} is solely applicable for architecture design within CNN search spaces, and LayerNAS \cite{15} is exclusively suitable for hierarchically ordered search spaces. 
These limitations impede the application of MP-NAS methods to advanced search spaces, such as SuperTransformer \cite{13,5}.
Besides, alternative strategies for effective performance evaluation of candidate networks are also expected, despite the improvement brought by parameter sharing \cite{52,54}, performance prediction based on learning curves \cite{55,56} and so on.

In this study, we introduce MathNAS, a novel MP-NAS framework for universal network architecture search. 
In contrast to previous studies which estimate the performance of a network by solving a whole problem, MathNAS adopts an alternative divide-and-conquer approach. 
In brief, MathNAS improves the performance evaluation of a candidate network by estimating the performance of each block of the network first, and then combining them to predict the overall performance of the network. 
Although estimating block performance involves network training, predicting network performance is completely training-free. 
Therefore, this approach reduces the complexity of network performance evaluation.
MathNAS achieves further improvement of the search strategy by transforming NAS to a programming problem, reducing the search complexity to polynomial time.

MathNAS contains three key steps:
\begin{itemize}
    
    \item \textbf {Block performance estimation:} The performance of each block is estimated by the performance difference between networks having and having not that specific block.
    
    \item \textbf {Network performance prediction:} The performance of a network is predicted based on the performances of its blocks.
    
    \item \textbf {NAS by ILP:} Utilizing the block performances, NAS is solved as an Integer Linear Programming (ILP) problem.
    
\end{itemize}

We perform experiments on search spaces with various network architectures, including NAS-Bench-201 (GNN) \cite{16}, MobileNetV3 (CNN) \cite{17}, SuperTransformer (Transformer) \cite{5} and NASViT (CNN+Trans) \cite{13}. 
Our experiments demonstrate that predicting network performance based on its blocks' performances is applicable to different network architectures.
In particular, the Spearman coefficient between the actual and the predicted top-1 indices on four different search spaces achieve 0.97, 0.92, 0.93, and 0.95, respectively. 
At the same time, by using the merit of the divide-and-conquer strategy to transform NAS into an ILP problem, MathNAS can find models superior to state-of-the-art (SOTA) models across different search spaces and tasks. In CV tasks, MathNAS achieves 82.5\% top-1 accuracy on ImageNet-1k with 1.5G FLOPs and 15M parameters, outperforming AutoFormer-small (81.7\%) and LeViT-256 (81.6\%). In NLP tasks, MathNAS reaches a Blue Score of 28.8, on par with Transformer (28.4), but requiring only 1/5 of the FLOPs.
In summary, our contributions are as follows:
\begin{enumerate}

    \item We propose a general framework for performance evaluation of candidate networks by estimating block performance first and then combining them to predict network performance, which greatly improves the evaluation efficiency.
    
    \item By virtue of the established mapping between block performance and network performance, we transform NAS into an ILP problem, which reduces the search complexity to polynomial.

    \item We demonstrate MathNAS by considering three key performance indices for network design, i.e. accuracy, latency and energy, and achieve results superior to SOTA models.

\end{enumerate}

\section{The Proposed Method}
\label{sec31}
\paragraph{Search Space and Notations.}

In this paper, we consider a widely used modular search space $\mathcal{S}=\left\{\mathcal{N}_1,...,\mathcal{N}_k,...\right\}$, in which network $\mathcal{N}_k$ consists of $m$ block nodes $\mathcal{B}_i (1 \leq i \leq m)$.
For a block node, there is $n$ alternative blocks, i.e.  $\mathcal{B}_i=\left\{b_{i,1},...,b_{i,j},...,b_{i,n}\right\}$, where block $b_{i,j} (1 \leq i \leq m, 1 \leq j \leq n)$ represents the $j$-th implementation of the $i$-th block node.
A network can therefore be denoted as $\mathcal{N}=(b_{1,\mathcal{J}_1},...,b_{i,\mathcal{J}_i},...,b_{m,\mathcal{J}_m})$, with its $i$-th block node implemented by block $b_{i,\mathcal{J}_i}$.
Totally, there are $m^n$ possible networks in the search space.
Here, we focus on three key performance indices for network design, i.e. accuracy, latency and energy consumption.
The accuracy, latency and energy consumption of network $\mathcal{N}$ are denoted as $Acc(\mathcal{N})$, $Lat(\mathcal{N})$, $Eng(\mathcal{N})$
, respectively.

\subsection{Problem Formulation: Reduce NAS Search Complexity from  $\mathcal{O}(n^m)$ to $\mathcal{O}(m*n)$.}
\label{sec31}
The objective of hardware-aware NAS is to find the best neural architecture $\mathcal{N^*}$ with the highest accuracy under limited latency $\hat{L}$ and energy consumption $\hat{E}$ in the search space $\mathcal{S}$:
\begin{equation}  
     \mathcal{N^*} = \underset{\mathcal{N} \in \mathcal{S}}{\arg\max}  Acc(\mathcal{N}), 
     s.t. Lat(\mathcal{N}) \leq \hat{L} ,Eng(\mathcal{N}) \leq \hat{E}
\label{equa1}
\end{equation} 
In order to fulfill the above goal, NAS has to search a huge space, evaluate and compare the performance of candidate networks.  
Early NAS studies usually fully train the candidate networks to obtain their performance ranking, which is prohibitively time-consuming.
Subsequent works introduce various acceleration methods.
For instance, the candidate networks can avoid training from scratch by sharing weights \cite{52,54}.
Also, the performance of a candidate network can be predicted based on the learning curve obtained from early termination of an incomplete training \cite{55,56}.
Despite of these improvements, a candidate network has to be trained, no matter fully or partially, to obtain a reasonable performance evaluation. And the huge number of candidate networks in the search space poses a formidable efficiency challenge.

However, the modular nature of the search space may provide us with a novel possibility. Although there are $n^m$ candidate networks in the search space, they are all constructed by the $n*m$ blocks. If we can evaluate the performances of the blocks by training of a limited number of networks, and if we can combine these block performance indices to obtain a reasonable performance evaluation of networks, we can reduce the complexity of network performance evaluation from  $\mathcal{O}(n^m)$ to  $\mathcal{O}(n*m)$.

Guided by this idea, we reformulate the search of
$\mathcal{N}^{*}$ from $\mathcal{S}$ as a succession of sub-problems. Each sub-problem corresponds to the task of searching the block $b_{i,\mathcal{J}^{*}_i}$ with the highest accuracy within the block node $\mathcal{B}_{i}$. This approach notably simplifies the original problem:
\begin{equation} 
\begin{split}
	\mathcal{N^*} = ( b_{1,\mathcal{J}^*_1},b_{2,\mathcal{J}^*_2}, \dots &,b_{m,\mathcal{J}^*_m} ) = 
	\underset{b_{1,j} \in \mathcal{B}_1}{\arg\max}  (b^A_{1,j}) \oplus 
	\underset{b_{2,j} \in \mathcal{B}_2}{\arg\max}  (b^A_{2,j}) \oplus \dots \oplus
	\underset{b_{m,j} \in \mathcal{B}_m}{\arg\max}  (b^A_{m,j})
	\, , \, \\
	 &s.t.  \, \,\, \sum\nolimits_{i=1}^{m} b_{i,\mathcal{J}^*_i}^{L} \leq \hat{L} \, , \, \sum\nolimits_{i=1}^{m} b_{i,\mathcal{J}^*_i}^{E} \leq \hat{E},
\end{split}
\label{equ2}
\end{equation} 
where $b_{i,j}^{A}$, $b_{i,j}^{L}$, $b_{i,j}^{E}$ represent the accuracy, latency, and energy of block $b_{i,j}$ respectively. $\oplus$ denotes the operation of adding a block to the network. With this approach, each block $b_{i,j}$ is searched only once and the complexity can be effectively reduced to $\mathcal{O}(n*m)$.
However, due to the mutual influence between blocks, a unified understanding of the relationship between the performance of $\mathcal{N}$ and its constituent blocks remains elusive, posing a challenge to the application of this method.

\begin{figure*}[t]
 \centering
 \begin{minipage}{0.16\linewidth}
  
     \centerline{\includegraphics[width=\textwidth]{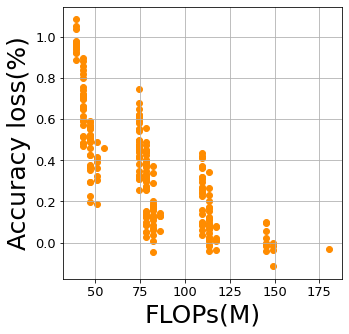}} 
 \end{minipage}
    \begin{minipage}{0.16\linewidth}
 
     \centerline{\includegraphics[width=\textwidth]{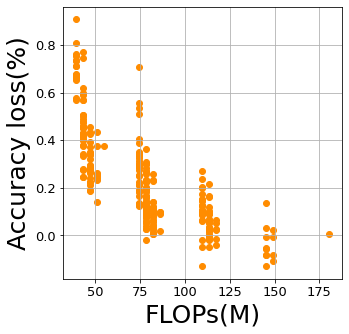}} 
 \end{minipage}
    \begin{minipage}{0.16\linewidth}
 
     \centerline{\includegraphics[width=\textwidth]{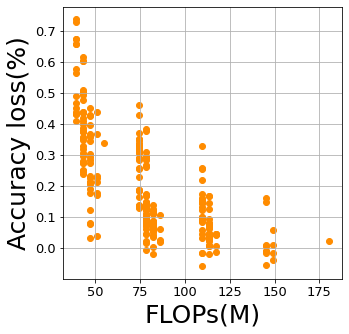}} 
   
 \end{minipage}
    \begin{minipage}{0.16\linewidth}
     \centerline{\includegraphics[width=\textwidth]{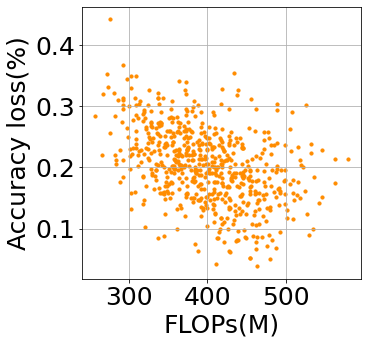}} 
 \end{minipage}
    \begin{minipage}{0.16\linewidth}
     \centerline{\includegraphics[width=\textwidth]{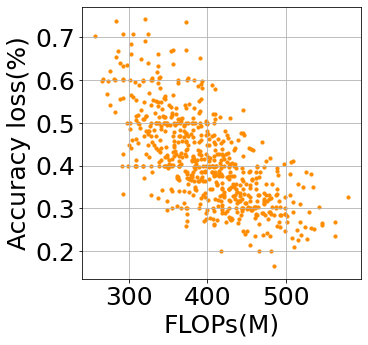}} 
 \end{minipage}
 \begin{minipage}{0.16\linewidth}
     \centerline{\includegraphics[width=\textwidth]{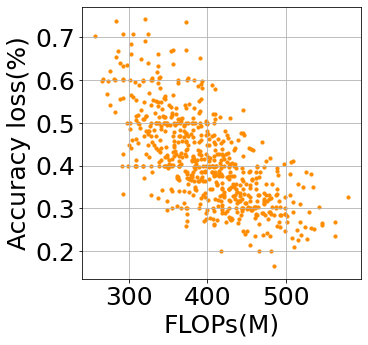}} 
 \end{minipage}
 \centerline{\small(a) $\Delta Acc(\mathcal{N}_{(i,1)\rightarrow (i,j)})$ vs $\mathcal{F}(\mathcal{N})$ on NAS-Bench-201 (left) and MobileNetV3 (right).}
   \begin{minipage}{0.16\linewidth}

     \centerline{\includegraphics[width=\textwidth]{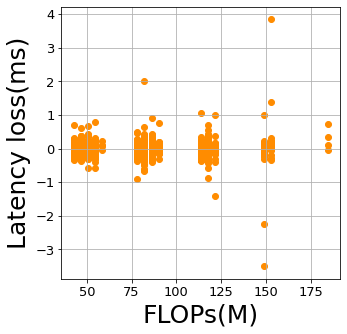}} 
 \end{minipage}
    \begin{minipage}{0.16\linewidth}
 
     \centerline{\includegraphics[width=\textwidth]{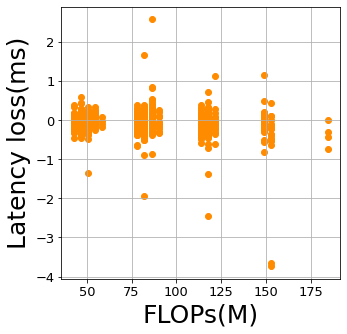}} 
 \end{minipage}
    \begin{minipage}{0.16\linewidth}
 
     \centerline{\includegraphics[width=\textwidth]{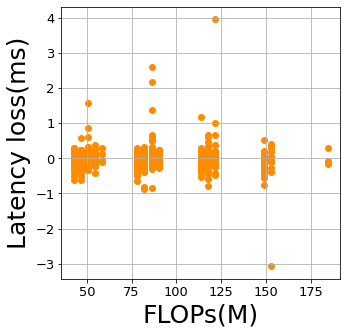}} 
   
 \end{minipage}
    \begin{minipage}{0.16\linewidth}
     \centerline{\includegraphics[width=\textwidth]{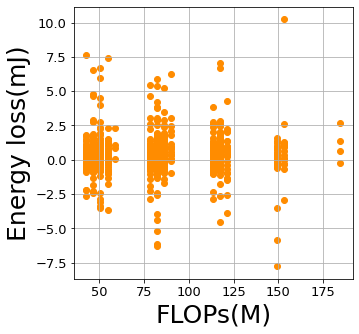}} 
 \end{minipage}
    \begin{minipage}{0.16\linewidth}
     \centerline{\includegraphics[width=\textwidth]{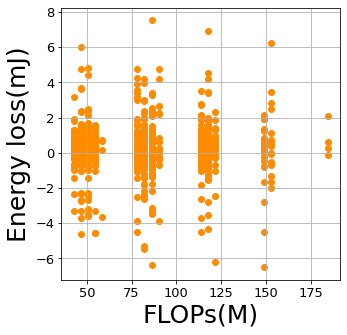}} 
 \end{minipage}
     \begin{minipage}{0.16\linewidth}
     \centerline{\includegraphics[width=\textwidth]{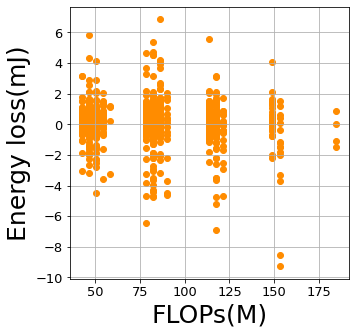}} 
 \end{minipage}
 \centerline{\small(b) $\Delta Lat(\mathcal{N}_{(i,1)\rightarrow (i,j)})$ (left)/ $\Delta Eng(\mathcal{N}_{(i,1)\rightarrow (i,j)})$ (right) vs $\mathcal{F}(\mathcal{N})$ in NAS-Bench-201 on edgegpu.}
     
	\caption{ 
	The relationship between $\Delta Acc(\mathcal{N}_{(i,1)\rightarrow (i,j)})$, $\Delta Lat(\mathcal{N}_{(i,1)\rightarrow (i,j)})$, $\Delta Eng (\mathcal{N}_{(i,1)\rightarrow (i,j)})$ and $\mathcal{F}(\mathcal{N})$.
	We conduct experiments on two search spaces: (1) NAS-Bench-201 \cite{16}. Graph Network. We use the accuracy obtained on ImageNet after training each network independently. We count the accuracy of all networks in the search space.
    (2) MobileNetv3 \cite{17}: sequentially connected CNN network. 
    We use the accuracy obtained by the network with shared weights. We sample 5 blocks per block node and count the accuracies of 3125 subnetworks on the ImageNet validation set.
	}
	\label{relationship}
\end{figure*} 

\subsection{Divide-and-Conquer: Network-Block-Network}
Consider the switching process $\mathcal{N}_{(i,1)\rightarrow (i,\mathcal{J}_i)}$, signifying the selection of the $i$-th $\mathcal{B}$ in network $\mathcal{N}$ as it switches from $b_{i,1}$ to $b_{i,\mathcal{J}_i}$, with other selected blocks remaining unchanged. Thus, any network $\mathcal{N} = ({b}_{1,\mathcal{J}_1}, {b}_{2,\mathcal{J}_2}, ...,{b}_{m,\mathcal{J}_m})$ can be viewed as the outcome of the base network $\widetilde{\mathcal{N}}=({b}_{1,1}, {b}_{2,1}, ...,b_{m,1})$ undergoing $m$ sequential switching processes.
Guided by this idea, we explore two aspects:
\label{sec32} 
\par
\textbf{E1: Can block performance be directly calculated as with network performance?}
\par
Considering the entire search space $\mathcal{S}$, let us denote the collection of networks with the selected block $b_{i,1}$ as $\scriptstyle\mathcal{N}^{\Omega}_{(i,1)}$, which comprises $n^{m-1}$ networks. For any network $\scriptstyle\mathcal{N}_{(i,1)}$ in $\scriptstyle\mathcal{N}^{\Omega}_{(i,1)}$, the switching process $\scriptstyle\mathcal{N}_{(i,1)\rightarrow (i,j)}$ signifies the selection of the $i$-th $\mathcal{B}$ in network $\mathcal{N}$ as it switches from $b_{i,1}$ to $b_{i,j}$, with other selected blocks remaining unchanged.  
In this switch, two changes occur. 
The first change sees the selected block in $\mathcal{B}_i$ switching from $b_{i,1}$ to $b_{i,j}$.The second change arises from an internal adjustment in $\mathcal{B}_i$, modifying its interactions with other block spaces in the network.
These changes lead to a difference in performance between $\scriptstyle\mathcal{N}_{(i,1)}$ and $\scriptstyle\mathcal{N}_{(i,j)}$, denote as $\Delta Acc(\mathcal{N}_{(i,1)\rightarrow(i,j)}), \Delta Lat(\mathcal{N}_{(i,1)\rightarrow(i,j)}), \Delta Eng(\mathcal{N}_{(i,1)\rightarrow(i,j)})$. 
By averaging performance difference obtained from all $n^{m-1}$ switching processes, $\scriptstyle\mathcal{N}^{\Omega}_{(i,1)\rightarrow (i,j)}$, we can derive two key parameters:
\begin{enumerate}
    \item  $\scriptstyle\Delta\phi(\mathcal{B}_{(i,1)\rightarrow (i,j)})$,
    the change in inherent capability of $\mathcal{B}_i$.
    \item $\scriptstyle\Delta\Phi(\mathcal{B}_{(i,1)\rightarrow (i,j)})$,
    the change in the interactive capability of $\mathcal{B}_i$ within $\mathcal{S}$.
\end{enumerate}
Accordingly, we define the performances of $b_{i,j}$ as:
\begin{equation} 
	b^{A}_{i,j}=\overline{\Delta Acc\left(\mathcal{N}^{\Omega}_{(i,1) \rightarrow  (i,j)}\right )}  = \Delta\phi(\mathcal{B}_{(i,1)\rightarrow (i,j)}) + \Delta\Phi(\mathcal{B}_{(i,1)\rightarrow (i,j)})
\label{equ3}
\end{equation} 
Similarly, $b_{i,j}^L$ and $b_{i,j}^E$ can be calculated employing the same methodology.
The unbiased empirical validation and theoretical proof supporting this method can be found in the Appendix.

\textbf{E2: How can we predict network performance using block performance?}
\par
To accurately derive the performance difference stemming from the switching process, we consider the performance difference that a particular block switch brings to different networks. We performed the identical process $\mathcal{N}_{(i,1)\rightarrow (i,\mathcal{J}_i)}$ over a range of networks within $\mathcal{N}^{\Omega}_{(i,1)}$. The outcome is illustrated in Figure 1, which depicts the relationship between the differences of three performances—latency, energy, and accuracy—and the FLOPs of the network. Our findings reveal that, within the same switching operation, latency and energy differences maintain consistency across different networks, while accuracy differences exhibit an inverse proportionality to the network's FLOPs, $\mathcal{F}(\mathcal{N})$.

This finding is logical, as networks with smaller FLOPs have lower computational complexity, rendering them more susceptible to block alterations. Conversely, networks with larger FLOPs exhibit higher computational complexity, making them less sensitive to individual block switching.
In the Appendix, we proceed to fit the inverse relationship between accuracy difference and network FLOPs using various formulas. The results suggest that the accuracy difference is approximately inversely related to the network's FLOPs. Consequently, we can posit $\Delta Acc(\mathcal{N}_{(i,1)\rightarrow (i,\mathcal{J}_i)}) = \alpha*1/\mathcal{F}(\mathcal{N})$, where $\alpha$ represents a specific block's coefficient.

\begin{figure*}[t]
	\centering
	\begin{minipage}{0.19\linewidth}
		\vspace{3pt}
		\centerline{\includegraphics[width=\textwidth]{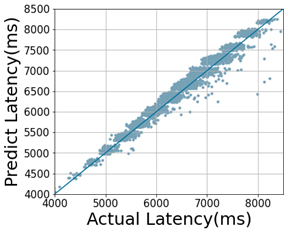}}
	\end{minipage}
	\hspace{10pt}
	\begin{minipage}{0.19\linewidth}
		\vspace{3pt}
		\centerline{\includegraphics[width=\textwidth]{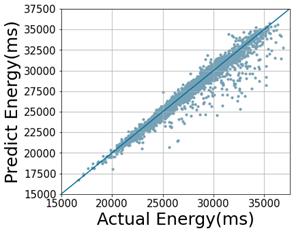}} 
	\end{minipage}
	\hspace{10pt}
	\begin{minipage}{0.19\linewidth}
		\vspace{3pt}
		\centerline{\includegraphics[width=\textwidth]{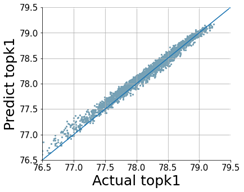}} 
	\end{minipage}
	\hspace{10pt}
	\begin{minipage}{0.19\linewidth}
		\vspace{3pt}
		\centerline{\includegraphics[width=\textwidth]{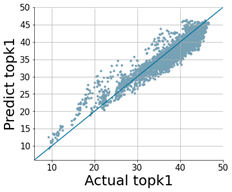}} 
	\end{minipage}
	 
	 \begin{minipage}{0.19\linewidth}
 
		\centerline{\includegraphics[width=\textwidth]{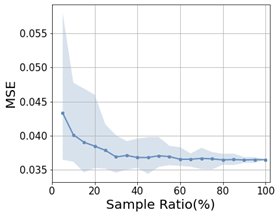}}
        \centerline{\small(a)Lat on NB201}
	\end{minipage}
	\hspace{10pt}
	\begin{minipage}{0.19\linewidth}
 
		\centerline{\includegraphics[width=\textwidth]{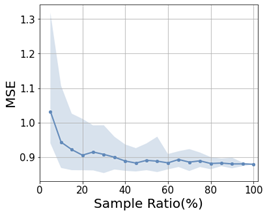}} 
		\centerline{\small(b)Energy on NB201}
	\end{minipage}
	\hspace{10pt}
	\begin{minipage}{0.19\linewidth}
 
		\centerline{\includegraphics[width=\textwidth]{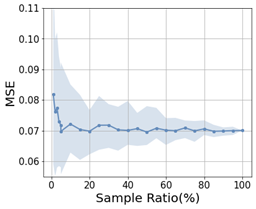}} 
		\centerline{\small(c)Acc on MBV3}
	\end{minipage}
	\hspace{10pt}
	\begin{minipage}{0.19\linewidth}
		\centerline{\includegraphics[width=\textwidth]{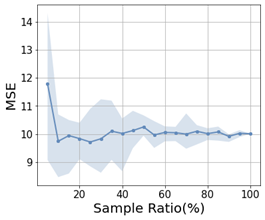}} 
	    \centerline{\small(d)Acc on NB201}
	\end{minipage}
	\caption{Validation of the predictive performance of Equation \ref{equa5} and the effect of sampling rate on it. The setup of the search space is the same as in Figure \ref{relationship}. 
	The first row shows the predictive performance of Equation \ref{equa5} when calculating $b_{i,j}^A$, $b_{i,j}^L$, and $b_{i,j}^E$ with Equation \ref{equ3}. 
	The second row shows the predictive performance of Equation \ref{equa5} when randomly sampling the network and computing the average difference to estimate $b_{i,j}^A$, $b_{i,j}^L$, and $b_{i,j}^E$ for the corresponding dataset.}
	\label{valide}
\end{figure*}

Based on these observations, we can deduce that for any network $\mathcal{N}_{(i,1)}$ within the set  $\mathcal{N}^{\Omega}_{(i,1)}$, the performance difference resulting from the switching process can be approximated as follows:
 \begin{equation} 
 \Delta Lat(\mathcal{N}_{(i,1)\rightarrow (i,j)})\approx b_{i,j}^L,
 \Delta Eng(\mathcal{N}_{(i,1)\rightarrow (i,j)})\approx b_{i,j}^E,
 \Delta Acc(\mathcal{N}_{(i,1)\rightarrow (i,j)})\approx b_{i,j}^A*\frac{\scriptstyle\overline{\mathcal{F}\left(\mathcal{N}^{\Omega}_{(i,j)}\right)}}{\scriptstyle \mathcal{F}(\mathcal{N}_{(i,j)})}
 \label{equ4}
 \end{equation}
By integrating Equation \ref{equ3} and \ref{equ4}, we can estimate the performance of $\mathcal{N} = ({b}_{1,\mathcal{J}_1}, {b}_{2,\mathcal{J}_2}, ...,{b}_{m,\mathcal{J}_m})$:
\begin{equation} 
\begin{split}
	Lat(\mathcal{N}) &= Lat(\widetilde{\mathcal{N}}) - \sum\nolimits_{i=1}^m{b_{i,\mathcal{J}_i}^L} \, \, , \,
    Eng(\mathcal{N}) = Eng(\widetilde{\mathcal{N}}) - \sum\nolimits_{i=1}^m{b_{i,\mathcal{J}_i}^E} \\
	Acc(\mathcal{N}) &= Acc(\widetilde{\mathcal{N}}) - \sum\nolimits_{i=1}^mb_{i,\mathcal{J}_i}^A*\cfrac{\scriptstyle\overline{\mathcal{F}\left( \mathcal{N}^{\Omega}_{(i,\mathcal{J}_i)}\right)}}{\scriptstyle\mathcal{F}(\mathcal{N}_{(i,\mathcal{J}_i)})}
	\label{equa5}
\end{split}
\end{equation}
\textbf{Proof-of-Concept Experiment.}
In our examination of the predictive efficacy of Equation \ref{equa5} across diverse network types, as depicted in Figure \ref{valide}, we observe that it accurately forecasts three performances for both sequentially connected networks and graph networks and both weight-independent and weight-shared networks, all without necessitating network training.
To the best of our knowledge, our method is the first work to precisely estimate network performances using a linear formula, and notably, it is theoretically applicable to all architectures.

\begin{wrapfigure}{r}{0.5\textwidth}
\vspace{-15pt}
  \begin{minipage}{0.5\textwidth}
    \begin{algorithm}[H]
       \SetAlgoHangIndent{0pt}
       \fontsize{8.0pt}{9pt}\selectfont
\caption{Math Neural Architecture Search}
\textbf{Stage1: Determine the Average-FLOPs Network}\\
\For{i=1,2,...,m}{ 
Calculate the average of the FLOPs of all blocks at the $i$-th note, $\overline{\mathcal{F}(b_i)}=(\mathcal{F}(b_{i,1})+...+\mathcal{F}(b_{i,n}))/n$;
\par
Select $b_{i,\mathcal{J}_i}$ whose FLOPs is closest to $\overline{\mathcal{F}(b_i)}$.\\
}
Define average-FLOPs net $\mathcal{N}^{avg}=\{b_{1,\mathcal{J}_1},..,b_{m,\mathcal{J}_m}\}$.
\par
\textbf{Stage2: Calculate Block Performances}\\  
\For{i=1,2,...,m}{
\For{j=1,2,...,n}{
Switch the $i$-th block in $\mathcal{N}^{avg}$ from $b_{i,1}$ to $b_{i,j}$;
\par
Calculate the performance difference of the network brought about by switching and use it as the performance of the block $b_{i,j}^A, b_{i,j}^L, b_{i,j}^E$. 
}
}
\par
\textbf{Stage3: Prediction and Architecture Search}\\
Calculate three characteristics of the base net $\widetilde{\mathcal{N}}=\{b_{1,1},\dots,b_{m,1}\}$ as $Acc(\mathcal{\widetilde{N}}),Lat(\mathcal{\widetilde{N}}),Eng(\mathcal{\widetilde{N}})$.
\par
For network $\mathcal{N}= \{b_{1,\mathcal{J}_1},b_{2,\mathcal{J}_2},...,b_{m,\mathcal{J}_m}\}$, its accuracy, latency and energy can be estimated by Equa. \ref{equa5}.
\par
Set required accuracy/latency/energy limit, and solve the corresponding ILP problem to obtain the architecture.
\label{arg1} 
\end{algorithm}

        \vspace{-30pt}
    \end{minipage}
\end{wrapfigure}

\subsection{Simplification: Single-sampling Strategy Instead of Full-sampling}
\label{sec33}
Despite the outstanding predictive performance displayed by Equation \ref{equa5}, its computation of loss averages proves to be costly. 
In this section, by employing a single-sample sampling strategy, we effectively reduce the time complexity from $O(n^m)$ to $O(n*m)$, enhancing efficiency without compromising precision.

\paragraph{Partial Network Sampling Strategies.}
We begin by investigating the sample size requirements for Equation \ref{equa5}. The second row of Figure \ref{valide} demonstrates rapid convergence of Equation \ref{equa5} with a notably small sample count for performance prediction. Specifically, Figure \ref{valide}(c)(d) reveals that a mere 5\% random sampling of networks is sufficient for the prediction to converge towards optimal performance. This underscores the impressive efficacy of Equation \ref{equa5}, which exhibits rapid convergence even with a limited number of random samples.

\begin{figure*}[t]
	\centering
	\begin{minipage}{0.23\linewidth}
		\centerline{\includegraphics[width=\textwidth]{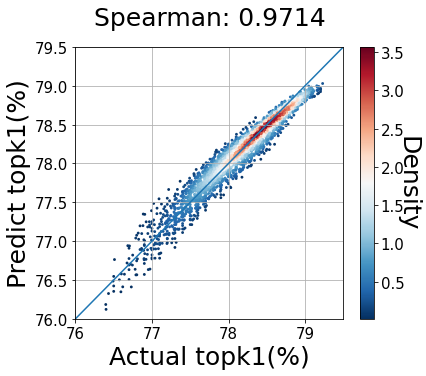}}
         \centerline{\small(a)Acc on MBv3}
         
	\end{minipage}
	\begin{minipage}{0.23\linewidth}
 
		\centerline{\includegraphics[width=\textwidth]{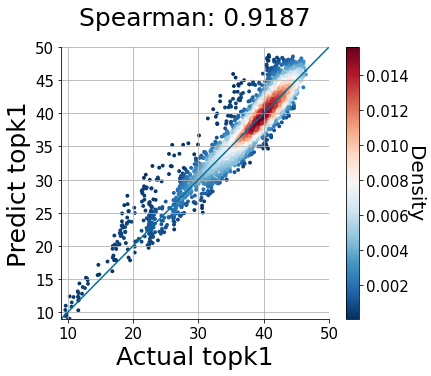}} 
		 \centerline{\small(b)Acc on NB201}
	\end{minipage}
	\begin{minipage}{0.23\linewidth}
 
		\centerline{\includegraphics[width=\textwidth]{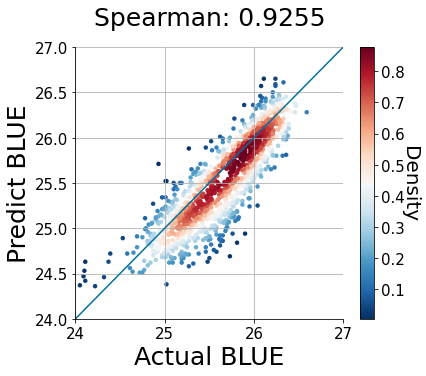}} 
		 \centerline{\small(c)Acc on SuperTran}
	\end{minipage}
	\begin{minipage}{0.23\linewidth}
 
		\centerline{\includegraphics[width=\textwidth]{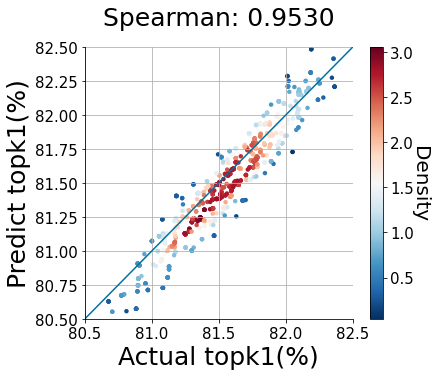}} 
		 \centerline{\small(d)Acc on SuperViT}
	\end{minipage}
	
	\begin{minipage}{0.23\linewidth}
 
		\centerline{\includegraphics[width=\textwidth]{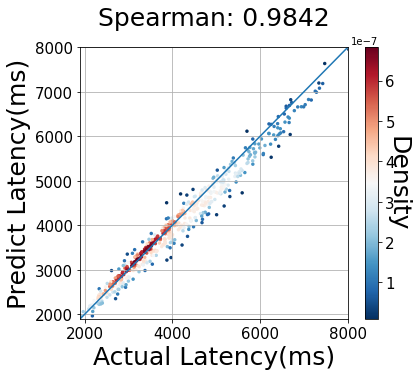}} 
		 \centerline{\small(e)Lat on MBv3}
	\end{minipage}
	\begin{minipage}{0.23\linewidth}
		\centerline{\includegraphics[width=\textwidth]{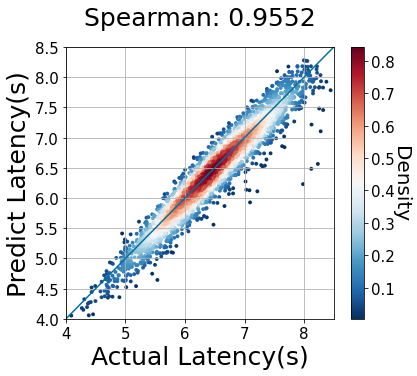}} \centerline{\small(f)Lat on NB201}
	 
	\end{minipage}
	\begin{minipage}{0.23\linewidth}
 
		\centerline{\includegraphics[width=\textwidth]{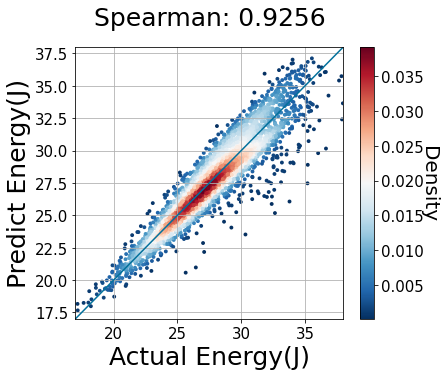}} 
		 \centerline{\small(g)Ener on NB201}
	\end{minipage}
	\begin{minipage}{0.23\linewidth}
		\centerline{\includegraphics[width=\textwidth]{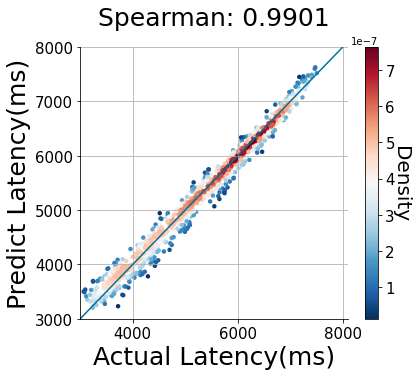}} 
		 \centerline{\small(h)Lat on SuperTran}
	\end{minipage}
 
	\caption{MathNAS algorithm verification. We conduct experiments on 4 search space. (1)MobileNetV3 \cite{17} (2)NAS-Bench-201 \cite{16} (3)SuperTransformer \cite{5} (4)SuperViT \cite{13}. 
	For NAS-Bench-201, we use the accuracy of each network trained individually as $Acc(\mathcal{N})$. For other spaces, the validate accuracy under shared weights base net is used.
	We show accuracy predictions on these four networks as well as hardware efficiency predictions on them. The calculation of $b_{i,j}^A, b_{i,j}^L$ and $b_{i,j}^E$ follows Algorithm \ref{arg1}. For NAS-Bench-201, we verify all nets and other spaces, we randomly sample 1000 nets to verify the prediction effect.}
	\label{argovalide}
\end{figure*}

\paragraph{Single Network Sampling Strategy.}
Building upon Equation \ref{equ4}, we can select an average-FLOPs network, denoted as $\mathcal{N}^{avg}$, in the search space with FLOPs approximating the mean value, ensuring ${\scriptstyle\overline{\mathcal{F}\left(\mathcal{N}^{\Omega}_{(i,j)}\right)}}/{\scriptstyle \mathcal{F}(\mathcal{N}^{avg}_{(i,j)})} \approx 1$. This leads to:
$b_{i,j}^A  \approx  \Delta Acc(\mathcal{N}^{avg}_{(i,1)\rightarrow (i,j)})\,,\,
	b_{i,j}^L  \approx  \Delta Lat(\mathcal{N}^{avg}_{(i,1)\rightarrow (i,j)})\,,\,
	b_{i,j}^E  \approx  \Delta Eng(\mathcal{N}^{avg}_{(i,1)\rightarrow (i,j)})$
By incorporating $\mathcal{N}^{avg}$ into Equation \ref{equa5}, we gain the ability to calculate any network performance.
Thus, we only need to verify the performance of $\mathcal{N}^{avg}_{(i,1)}$ and $\mathcal{N}^{avg}_{(i,j)}$ for $b_{i,j}$, resulting in an $O(n*m)$ search time complexity. The complete MathNAS algorithm is presented in Algorithm \ref{arg1}, which, in theory, can achieve accurate prediction on any network structure with polynomial search complexity.

\paragraph{Validation experiment of single-sample strategy effectiveness.}
We assess the efficacy of Algorithm \ref{arg1} across CNN, GCN, Transformer, and CNN+Transformer search spaces, with the outcomes displayed in Figure \ref{argovalide}. It is evident that the Spearman correlation coefficient between predicted and actual values exceeds 0.9 for all networks. Remarkably, accuracies of 0.95 and 0.97 are attained on ViT and CNN architectures, respectively, emphasizing the algorithm's robustness.

\subsection{MathNAS: Converting Architecture Search to ILP.}
\label{sec34}
We denote $b_{i,j}^F$ as the FLOPs of block $b_{i,j}$, and $b_{i,j}^B\in \left\{0,1\right\}$ as the indicator of whether block $b_{i,j}$ is used in a network. If block $b_{i,j}$ is selected as the implementation of block node $\mathcal{B}_i$ in a network, $b_{i,j}^B=1$, otherwise 0.
The problem that NAS needs to solve can be formulated as:
\begin{equation}
\begin{split}
     &\max \limits_{b^B}Acc(\widetilde{\mathcal{N}}) - \cfrac{\sum_{i=1}^m\sum_{j=1}^nb_{i,j}^A*b_{i,j}^B}{\sum_{i=1}^m\sum_{j=1}^nb_{i,j}^F*b_{i,j}^B}*\overline{\mathcal{F}(\mathcal{N})}\\
     s.t.\,\,\,
    &Lat(\widetilde{\mathcal{N}})-\sum_{i=1}^m\sum_{j=1}^nb_{i,j}^L*b_{i,j}^B\leq \hat{L},
    Eng(\widetilde{\mathcal{N}})-\sum_{i=1}^m\sum_{j=1}^nb_{i,j}^E*b_{i,j}^B\leq \hat{E},\\
    &\sum\nolimits_{j=1}^nb_{i,j}^B=1, b_{i,j}^B\in\left\{0,1\right\}, \forall{1\leq i \leq m}.
\end{split}
\label{eq14}
\end{equation}
The objective is to obtain the maximum accuracy network under two constraints.
First, the latency and energy cannot exceed the limit.
Second, for any block node, only one block is used.
As Equation \ref{eq14} is a fractional objective function, it can be transformed into an ILP problem by variable substitution.

\begin{figure}[b]
	\centering
	\begin{minipage}{0.23\linewidth}
		\centerline{\includegraphics[width=\textwidth]{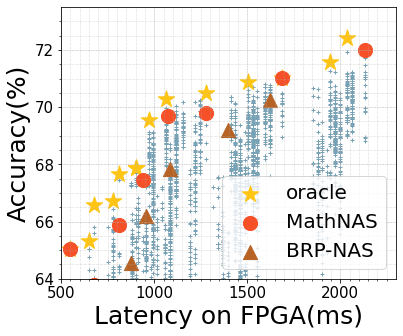}}
	\end{minipage}
	\begin{minipage}{0.24\linewidth}
		\centerline{\includegraphics[width=\textwidth]{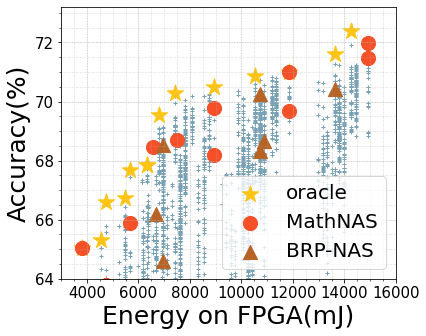}}
	\end{minipage}
	\hspace{5pt}
	\begin{minipage}{0.225\linewidth}
		\centerline{\includegraphics[width=\textwidth]{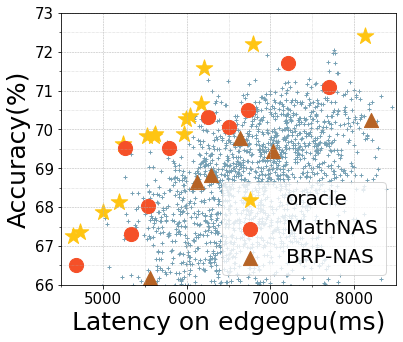}}
	\end{minipage}
	\begin{minipage}{0.225\linewidth}
		\centerline{\includegraphics[width=\textwidth]{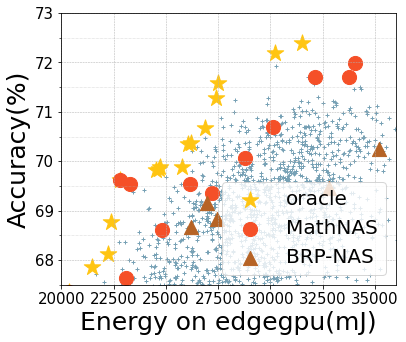}}
	\end{minipage}
	
	\caption{A comparison of networks searched by MathNAS (red circles) versus those searched by BRP-NAS (brown triangles) in the NAS-Bench-201 space. Across different devices, the networks searched by MathNAS demonstrate a closer proximity to the Pareto front (yellow five-pointed stars) as compared to the networks obtained through BRP-NAS.}
	\label{nas-bench-parato}
\end{figure}

\section{Performance Evaluation}
\label{sec5}
Experiments were conducted at three levels to evaluate the efficacy of MathNAS. 
Firstly, we validated the effectiveness of MathNAS on CV tasks by conducting searches across three different search spaces. 
Then we employed MathNAS to design efficient architectures for NLP tasks, showcasing its remarkable generalization capabilities. 
Finally, we leveraged MathNAS to perform real-time searches on edge devices, considering hardware resources, and achieved exceptional on-device performance.

\subsection{Experimental Setup}
\textbf{Search space.}
For CV tasks, we validate our method on three search spaces:
(1) NAS-Bench-201 \cite{16}: a search space encompasses 15,625 architectures in a DARTS-like configuration.
(2) SuperViT \cite{13}: a hybrid search space that combines ViT and CNN, containing approximately $4 \times 10^{10}$ network architectures.
(3) MobileNetV3 \cite{17}: a lightweight network search space comprising about $10^{10}$ network architectures.
For NLP tasks, we validate our approach on the SuperTransformer search space \cite{5}, which includes $10^{15}$ networks within a lightweight Transformer framework.

\textbf{Search and training settings.}
For NAS-Bench-201 and MobileNetV3, we adopt the training methodology employed in \cite{52} and \cite{32} to train the base net for 100 epochs. 
Subsequently, we conducted MathNAS search on the base net. 
As for SuperTransformer and SuperViT, we adhere to the training algorithm proposed by \cite{13} to train the base net for 100 epochs before conducting MathNAS search.
The settings of hyperparameters in the training are consistent with the original paper.
We employ the Gurobipy solver to address the ILP problem. In the SuperViT and SuperTransformer search spaces, we impose a search time limit of 10 seconds to expedite the process. 
For the other search spaces, we do not enforce any time constraints.

The search cost of MathNAS consists of two stages: offline network pre-training that is conducted only once and online real-time search.
During the offline network pre-training, MathNAS evaluates block performance once.
During online searching, MathNAS is capable of multiple real-time searches based on the current hardware resource constraints.
To negate the influence of variations in GPU models and versions on the pre-training time, and to facilitate comparisons by future researchers, we have adopted pre-trained networks provided by existing works. All mentions of search cost in the paper refer solely to the real-time search time on edge devices.
We provide a detailed description of the search space, more experimental results, and visualizations of the searched architectures in the Appendix.

\begin{figure}[t]
\begin{minipage}{.5\linewidth}
\centering

\includegraphics[width=1\textwidth]{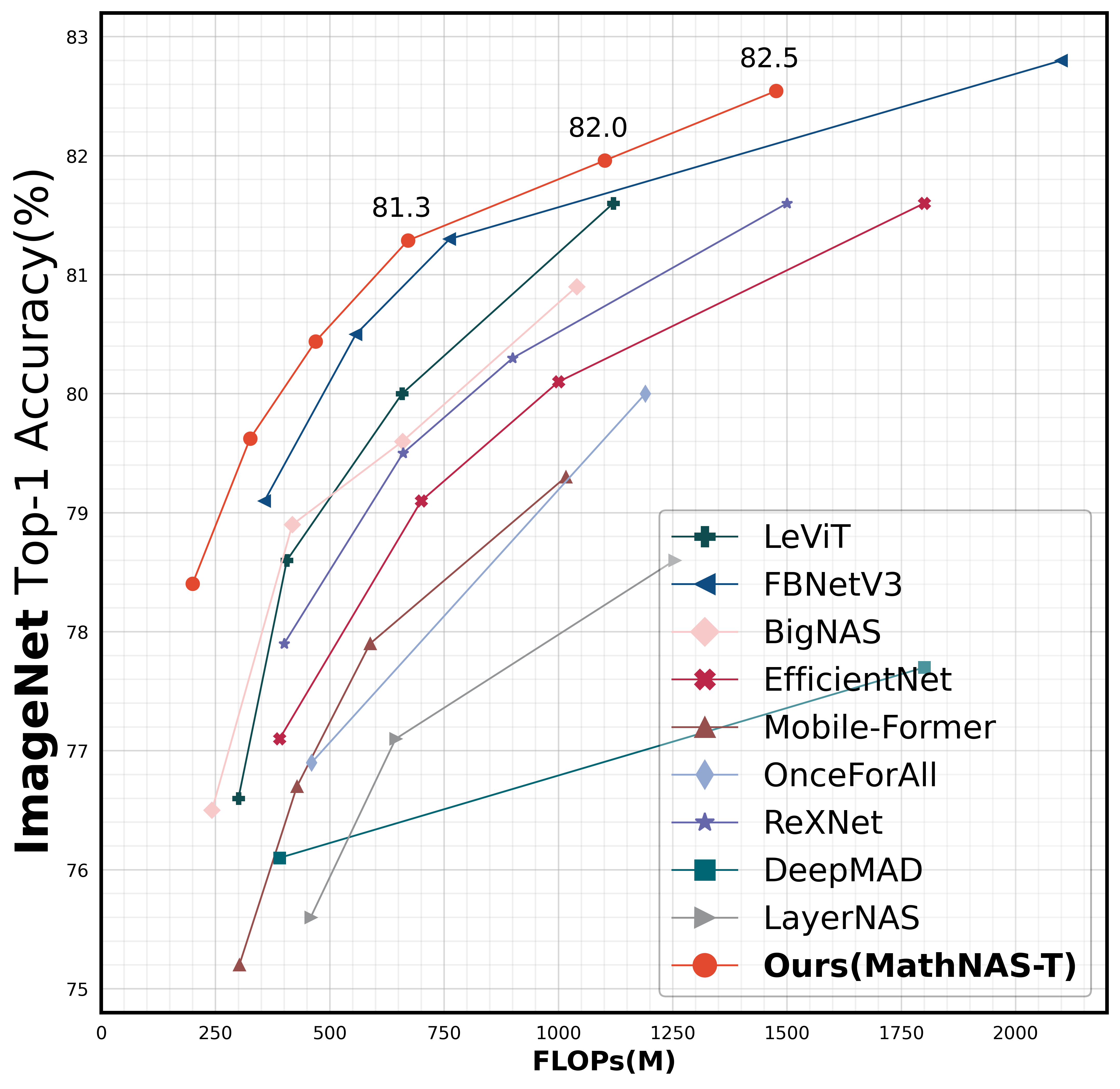}

\label{fig3}
\end{minipage}\quad
\begin{minipage}{.45\linewidth}

\centering
 \hspace{-20pt}
 \vspace{10pt}
\resizebox{1\linewidth}{!}{
\begin{tabular}{cccccc}
		\toprule
		 Model&Top-1(\%)&Top-5(\%)&$\mathcal{F}$(M)&Param(M)\\
		 \midrule
		 \midrule
		 ResNet-18\cite{23}&77.7&93.2&1800&11.7\\ 
		 ReXNet\cite{24}&77.9&93.9&400&4.8\\
		 \cdashline{1-5}[2pt/2pt]
		 \rule{0pt}{10pt}
		 \textbf{MathNAS-T1}&\textbf{78.4}&\textbf{93.5}&\textbf{200}&\textbf{8.9}\\
		 \midrule
		 LeViT-128\cite{25}&78.6&94.0&406&9.2\\
		 EfficientNet-B1\cite{26}&79.1&94.4&700&7.8\\ 
		 ReXNet\cite{24}&79.5&94.7&660&7.6\\
		 \cdashline{1-5}[2pt/2pt]
		 \rule{0pt}{10pt}
		 \textbf{MathNAS-T2}&\textbf{79.6}&\textbf{94.3}&\textbf{325}&\textbf{9.3}\\
		 
		 \midrule
		 ResNetY-4G\cite{27}&80.0&94.8&4000&21\\
		 LeViT-192\cite{25}&80.0&94.7&658&10.9\\
		 EfficientNet-B2\cite{26}&80.1&94.9&1000&9.2\\
		 ReXNet\cite{24}&80.3&95.2&900&9.7\\
		 ResNet-50\cite{23}&80.6&95.1&4100&25.6\\ 
		 \cdashline{1-5}[2pt/2pt]
		 \rule{0pt}{10pt}
		 \textbf{MathNAS-T3}&\textbf{81.3}&\textbf{95.1}&\textbf{671}&\textbf{13.6}\\
 
		 \midrule
		 Swin-T\cite{28}&81.3&95.5&4500&29\\
		 EfficientNet-B3\cite{26}&81.6&95.7&1800&12\\
		 LeViT-256\cite{25}&81.6&95.4&1120&18.9\\
		 ReXNet\cite{24}&81.6&95.7&1500&16\\
		 AutoFormer-small\cite{29}&81.7&95.7&5100&22.9\\
		  \cdashline{1-5}[2pt/2pt]
		 \rule{0pt}{10pt}
		 \textbf{MathNAS-T4}&\textbf{82.0}&\textbf{95.7}&\textbf{1101}&\textbf{14.4}\\
 
		 \midrule
		 AutoFormer-base\cite{29}&82.4&95.7&11000&54\\
		 LeViT-384\cite{25}&82.6&96.0&2353&39.1\\
		  
		  \cdashline{1-5}[2pt/2pt]
		 \rule{0pt}{10pt}
		 \textbf{MathNAS-T5}&\textbf{82.5}&\textbf{95.8}&\textbf{1476}&\textbf{14.8}\\

		\bottomrule
	\end{tabular}}
	
\end{minipage}
\caption{MathNAS v.s. SOTA ViT and CNN models on
ImageNet-1K.}
   
	\label{vitcompare}
\end{figure}

\subsection{MathNAS for Designing Effective CV Networks}
\paragraph{MathNAS for NAS-Bench-201.}
To assess the effectiveness of our method in striking a balance between accuracy and hardware efficiency, we compare networks searched by MathNAS under hardware efficiency constraints to those searched by BRP-NAS \cite{53}, which utilizes GNN predictors to estimate network performance. As illustrated in Figure \ref{nas-bench-parato}, MathNAS consistently locates networks that approach Pareto optimality in the majority of cases, whereas the employment of GNN predictors leads to suboptimal model choices. An extensive comparison between the searched architectures and SOTA models is provided in the Appendix for further insight.

\paragraph{MathNAS for ViT.}
To assess the performance of MathNAS in designing larger models, we utilize it to create effective ViT models for ImageNet-1K classification. Figure \ref{vitcompare} demonstrates that MathNAS surpasses or matches the performance of existing SOTA models. For instance, MathNAS-T5 achieves an accuracy of 82.5\%, which is comparable to LeViT-384 \cite{25} and Autoformer-base \cite{29}, while consuming only about 50\% and 15\% of their respective FLOPs. Similarly, MathNAS-T3 achieves comparable accuracy to RexNet \cite{24} but with approximately half the FLOPs.
MathNAS also exhibits exceptional performance in the realm of small networks. Particularly, MathNAS-T1 achieves a top-1 accuracy of 78.4\%, surpassing ResNet-18 \cite{23} and ReXNet \cite{24} by 0.7\% and 0.5\% respectively.

\begin{wraptable}{r}{0.5\linewidth}
\vspace{-15pt}
  \caption{Performance of mobile networks designed with MathNAS. Top-1 accuracy on ImageNet-1K.}
	\centering
	\resizebox{1\linewidth}{!}{
 
  \begin{tabular}{ccccc}
		\toprule
 
		 Model&FLOPs(M)&Top-1&Search Time&Scale Up\\
		 \midrule
		 \midrule
		 FBNet-b\cite{30}&295&74.1&609h&1.9$\times$\\
		 AtomNAS-A\cite{31}&258&74.6&492h&2.3$\times$\\
		 OFA\cite{32}&301&74.6&120h&9.6$\times$\\
		 
		 \textbf{MathNAS-MB1}&\textbf{257}&\textbf{75.9}&\textbf{0.9s}&\textbf{4.6M}$\times$\\
		 
		 \cdashline{1-5}[2pt/2pt]
		 \rule{0pt}{15pt}
		 MnasNet-A1\cite{33}&312&75.2&40025h&1.0$\times$\\
		 ProxylessNAS-R\cite{34}&320&74.6&520h&76.9$\times$\\
		 AtomNAS-B\cite{31}&326&75.5&492h&81.4$\times$\\
		 FairNAS-C\cite{20}&321&71.1&384h&104.2$\times$\\
		 Single Path One-Shot\cite{35}&323&74.4&288h&138.9$\times$\\
		 OFA\cite{32}&349&75.8&120h&333.5$\times$\\
		 
		 \textbf{MathNAS-MB2}&\textbf{289}&\textbf{76.4}&\textbf{1.2s}&\textbf{144M}$\times$\\
		 \cdashline{1-5}[2pt/2pt]
		 \rule{0pt}{15pt}
	 
		 EfficientNet B0\cite{4}&390&76.3&72000h&1.0$\times$ \\
		 FBNet-c\cite{30}&375&74.9&580h&124.1$\times$\\
		 ProxylessNAS-GPU\cite{34}&465&75.1&516h&139.5$\times$\\
		 AtomNAS-C\cite{31}&363&76.3&492h&146.3$\times$\\
		 FairNAS-A\cite{20}&388&75.3&384h&104.2$\times$\\
		 FairNAS-B\cite{20}&345&75.1&384h&104.2$\times$\\
		 
		 \textbf{MathNAS-MB3}&\textbf{336}&\textbf{78.2}&\textbf{1.5s}&\textbf{173M}$\times$\\
		 
		\cdashline{1-5}[2pt/2pt]
		 \rule{0pt}{15pt}
		 EfficientNet B1\cite{4}&700&79.1&72000&1.0$\times$ \\
		 MnasNetA1\cite{33}&532&75.4&40025&1.8$\times$\\
		 BigNAS-M\cite{36}&418&78.9&1152&62.5$\times$\\

		 \textbf{MathNAS-MB4}&\textbf{669}&\textbf{79.2}&\textbf{0.8s}&\textbf{324M}$\times$\\

		\bottomrule
	\end{tabular}}
	\vspace{-15pt}
	\label{mobileresult}
\end{wraptable}

\paragraph{MathNAS for Mobile CNNs.}
We employ MathNAS to design mobile CNN models for further investigation, conducting our search within the MobileNetV3 search space. As demonstrated in Table \ref{mobileresult}, MathNAS-MB4 achieves a top-1 accuracy of 79.2\%, which is on par with EfficientNet-B1 (79.1\%). It is important to note that EfficientNet-B1 is derived through a brute-force grid search, necessitating approximately 72,000 GPU hours \cite{4}. Despite this, MathNAS-MB4 offers comparable performance to EfficientNet-B1 while only requiring 0.8 seconds to solve an ILP problem on the GPU and search for a suitable network.
MathNAS also excels in the context of smaller networks. Notably, MathNAS-MB1 requires only 257M FLOPs to achieve a top-1 accuracy of 75.9\%, surpassing the performance of FBNet-b \cite{30}, AtomNAS-A \cite{31}, and OFA \cite{32}, all of which demand higher computational resources.

\subsection{MathNAS for Designing Effective NLP Networks}
We perform a comparative evaluation of MathNAS against SOTA NLP models on the WMT'14 En-De task to gauge its effectiveness. Table \ref{nlp} reveals that MathNAS surpasses all baseline models in terms of BLEU score while also achieving FLOPs reduction across three distinct devices. Specifically, on Intel Xeon CPUs, MathNAS with full precision attains a remarkable 74\% reduction in FLOPs compared to Transformer \cite{8} and a 23\% reduction compared to HAT \cite{5}, while registering improved BLEU scores by 0.4 and 0.3, respectively.
Additionally, MathNAS excels in designing lightweight NLP models. On Nvidia TITAN Xp GPUs under latency constraints, MathNAS yields FLOPs comparable to HAT \cite{5}, but with a 0.3 higher BLEU score. A noteworthy aspect is that the network search process facilitated by MathNAS requires only 10 seconds, considerably reducing search time. As a result, employing MathNAS leads to a reduction of over 99\% in $CO_2$ emissions compared to baseline models, underscoring its positive environmental impact.

\begin{table}[t]
\caption{MathNAS vs. SOTA baselines in terms of accuracy and efficiency on NLP tasks.}
	\centering
	\resizebox{\textwidth}{!}{
	\begin{tabular}{c|ccc|ccc|ccc|cc}
		\toprule
		 &\multicolumn{3}{|c}{Raspberry Pi}&\multicolumn{3}{|c}{Intel Xeon CPU}&\multicolumn{3}{|c}{Nvidia TITAN Xp GPU}&\multicolumn{2}{|c}{Search Cost}\\
		\midrule
		
		Model&FLOPs&BLEU&Latency&FLOPs&BLEU&Latency&FLOPs&BLEU&Latency&Time&$CO_2$\\
		\midrule
		\midrule
		Transformer\cite{8}&10.6G&28.4&20.5s&10.6G&28.4&808ms&10.6G&28.4&334ms&184h&52lbs\\
		Evolved Trans.\cite{37}&2.9G&28.2&7.6s&2.9G&28.2&300ms&2.9G&28.2&124ms&219200h&624000lbs\\
		HAT\cite{5}&1.5G&25.8&3.5s&1.9G&25.8&138ms&1.9G&25.6&57ms&200h&57lbs\\
		\textbf{MathNAS}&\textbf{1.7G}&\textbf{25.5}&\textbf{3.2s}&\textbf{1.8G}&\textbf{25.9}&\textbf{136ms}&\textbf{1.8G}&\textbf{25.9}&\textbf{68ms}&\textbf{10s}&\textbf{0.0008lbs}\\
		\midrule
		HAT\cite{5}& 2.3G&27.8&5.0s& 2.5G&27.9&279ms&2.5G&27.9&126ms&200h&57lbs\\
		\textbf{MathNAS}&\textbf{2.1G}&\textbf{28.3}&\textbf{4.7s}&\textbf{2.4G}&\textbf{28.6}&\textbf{272ms}&\textbf{2.0G}&\textbf{28.1}&\textbf{107ms}&\textbf{10s}&\textbf{0.0008lbs}\\
		\midrule
		HAT\cite{5}&3.0G&28.4&7.0s&3.5G&28.5&384ms&3.1G&28.5&208ms&200h&57lbs\\
		\textbf{MathNAS}&\textbf{2.8G}&\textbf{28.6}&\textbf{6.5s} &\textbf{2.8G}&\textbf{28.8}&\textbf{336ms}&\textbf{2.6G}&\textbf{28.6}&\textbf{189ms}&\textbf{10s}&\textbf{0.0008lbs}\\
 
		\bottomrule
	\end{tabular}}
	\label{nlp}
\end{table}

\subsection{MathNAS for Designing Dynamic Networks.}

\label{sec51}
Deployed on edge devices (Raspberry Pi 4b, Jetson TX2 CPU, TX2 GPU), MathNAS allows dynamic network switching suited to device conditions. 
Within the MobileNetV3 search space, we account for memory limitations by calculating performance indices for each block, subsequently deploying five selected Pareto-optimal blocks balancing accuracy and latency in each block node.
During runtime, latency is continuously monitored. Should it surpass a preset threshold, MathNAS immediately updates the blocks' latency. 
Then the device solves the ILP problem to identify the optimal network architecture, comparing and switching blocks with the searched network as required.

\begin{figure}[h]
	\centering
	\begin{minipage}{0.25\linewidth}
		\centerline{\includegraphics[width=\textwidth]{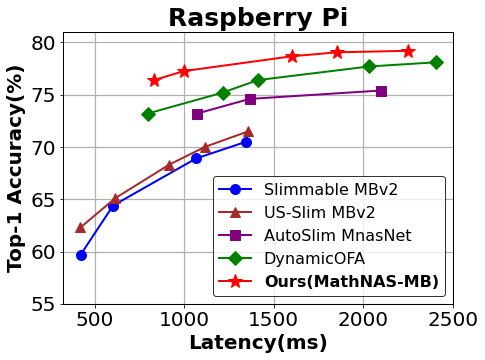}}
	\end{minipage}
	\begin{minipage}{0.25\linewidth}
		\centerline{\includegraphics[width=\textwidth]{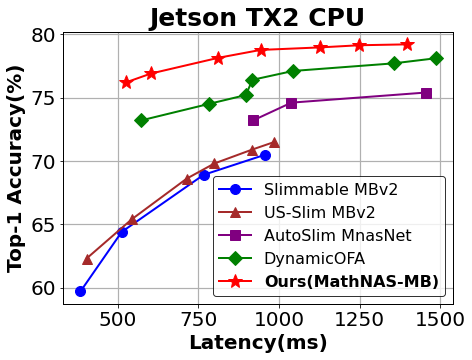}}
	\end{minipage}
	\hspace{5pt}
	\begin{minipage}{0.25\linewidth}
		\centerline{\includegraphics[width=\textwidth]{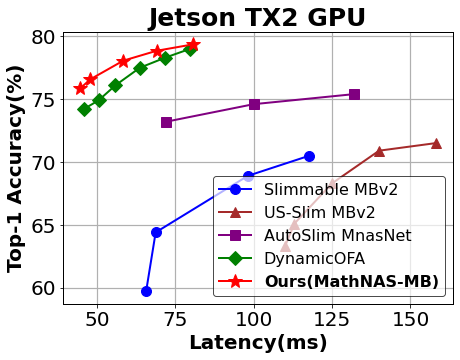}}
	\end{minipage}
	\begin{minipage}{0.18\linewidth}
		\centerline{\includegraphics[width=\textwidth]{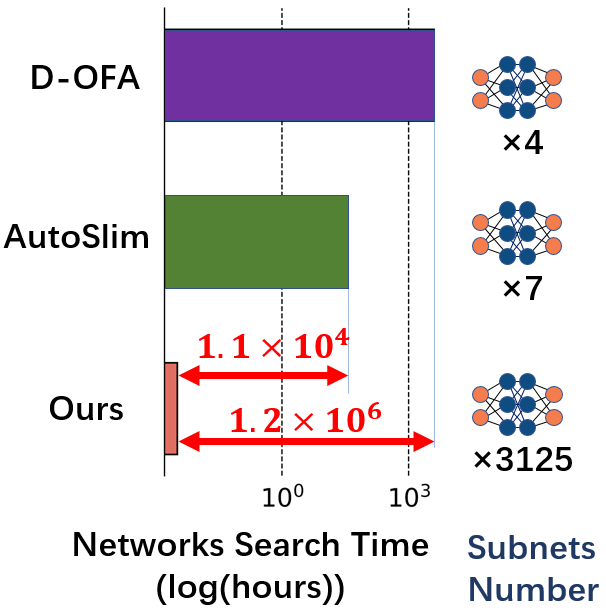}}
	\end{minipage}
	
	\caption{Top-1 vs. Latency of MathNAS over SOTA dynamic baselines on  three devices.}
	\label{dynamic-fig}
\end{figure}

In comparison with SOTA dynamic network models, MathNAS demonstrates superior performance as outlined in Table \ref{dynmictab} and Figure \ref{dynamic-fig}. Impressively, MathNAS solves the ILP problem on-device in a mere 0.4 seconds on the TX2 GPU, enabling real-time search. This notably enhances the number of executable networks on the device, outdoing SlimmableNet \cite{38} and USlimmableNet \cite{39} by factors of 781 and 116 respectively. Additionally, through a block-based approach, MathNAS enables efficient network alterations by replacing only essential blocks.
When compared to Dynamic-OFA \cite{41}, which shares similar performance, MathNAS significantly reduces the switching time by 80\%. The Appendix details the use of Pareto-optimal blocks and related network experiment results.

\begin{table}[h]
\caption{MathNAS vs. SOTA baselines in terms of Dynamic Networks.}
	\centering
	\resizebox{\textwidth}{!}{
	\begin{tabular}{c|cc|ccc|cc|cc|cc}
		\toprule
		&\multicolumn{2}{|c}{Network}&\multicolumn{3}{|c}{Latency (ms)}&\multicolumn{6}{|c}{On Device Performance}\\ 
		\midrule
		Model&\makecell{Top-1\\(\%)}&\makecell{FLOPs\\(M)}& \makecell{Raspb\\Pi}&\makecell{TX2\\CPU}& \makecell{TX2\\GPU}& \makecell{Search\\Method}&\makecell{Search\\Time}&\makecell{Switch\\Unit}&\makecell{Switch\\Time}&\makecell{Nets\\Number}&\makecell{Scale\\Up}\\
		\midrule
		\midrule
		S-MbNet-v2\cite{38} &70.5&301&1346&958&118&Manual Design&-&Channel&15ms&4&1.0x\\
		US-MbNet-v2\cite{39}&71.5&300&1358&959&158&Manual Design&-&Channel&18ms&27&6.7x\\ 
		AS-MNASNet\cite{40}&75.4&532&2097&1457&2097&Greedy Slimming&4000h&Channel&37ms&4&1.0x\\
		Dynamic-OFA\cite{41}&78.1&732&2404&1485&80&Random+Evplution&35h&Network&244ms&7&1.7x\\
		\cdashline{1-12}[2pt/2pt]
		 \rule{0pt}{20pt}
		\textbf{MathNAS}&\textbf{\makecell{75.9\\79.2}}&\textbf{\makecell{257\\669}}&\textbf{\makecell{832\\2253}}&\textbf{\makecell{525\\1398}}&\textbf{\makecell{76\\81}}&\textbf{On-Device Search}&\textbf{0.4-12s}&\textbf{Block}&\textbf{61ms}&\textbf{3125}&\textbf{781x}\\ 
		\bottomrule
	\end{tabular}}
	\label{dynmictab}
\end{table}

\section{Conclusion}
This paper introduces MathNAS, the first architecturally-general MP-NAS. 
By virtue of the modular nature of the search space, we introduce block performance and establish the mapping from block performance and network performance, which enables subsequent transformation of NAS to an ILP problem.
This novel strategy reduces network search complexity from exponential to polynomial levels while maintaining excellent network performance. 
We propose three transformation formulas to depict this process and support them with theoretical proofs and extensive experiments. 
MathNAS achieves SOTA performance on various large-scale CV and NLP benchmark datasets. 
When deployed on mobile devices, MathNAS enables real-time search and dynamic networks, surpassing baseline dynamic networks in on-device performance. 
Capable of conducting rapid searches on any architecture, MathNAS offers an appealing strategy for expediting the design process of large models, providing a clearer and more effective solution.

\section{Acknowledgment}
We thank the anonymous reviewers for their constructive comments.
This work was partially supported by National Key R\&D Program of China (2022YFB2902302). Sihai Zhang gratefully acknowledges the support of Innovative Research (CXCCTEC2200001).

\normalem
\bibliographystyle{unsrt}   
\bibliography{references}

\newpage

\appendix

\paragraph{Organization}
In this supplementary file, we provide in-depth descriptions of the materials that are not covered in the main paper, and report additional experimental results. The document is organized as follows:
\begin{itemize}
    \item \textbf{Section A}- Related work.
    \item \textbf{Section B}- Theoretical Analysis.
    \begin{itemize}
        \item \textbf{B.1} Unbiased proof.
        \item \textbf{B.2} Fit validation analysis.
        \item  \textbf{B.3} Ablation analysis.
        \item  \textbf{B.4} Effectiveness analysis.
        \item  \textbf{B.5} Space generality analysis.
        \item  \textbf{B.6} Searching equation solving details.
    \end{itemize}
    \item \textbf{Section C}- Experimental setups.    
    \item \textbf{Section D}- Additional experiments.
    \begin{itemize}
        \item \textbf{D.1} performance comparison on NAS-Bench-201.
        \item  \textbf{D.2} performance comparison with block-wise methods.
        \item  \textbf{D.3} Additional experiments on dynamic networks.
    \end{itemize}
    \item \textbf{Section E}- Searched Architecture Visualization.
    \begin{itemize}
        \item \textbf{E.1} Visualization Architecture on MobileNet-V3.
         \item  \textbf{E.2} Visualization Architecture on SuperViT Space.
         \item \textbf{E.3} Visualization Architecture on Dynamic Network.
         \end{itemize}
    \item \textbf{Section F}- Discussion.
    \begin{itemize}
        \item  \textbf{F.1} Limitation and Future Work.
         \item \textbf{F.2} Potential negative social impact.
    \end{itemize}
    \end{itemize}


\section{Related Work}
\label{sec3}
Neural Architecture Search (NAS) was introduced to ease the process of manually designing complex neural networks. Early NAS \cite{42} efforts employed a brute force approach by training candidate architectures and using their accuracy as a proxy for discovering superior designs.
Subsequent EA and RL-driven methods significantly enhanced search efficiency by sampling and training multiple candidate architectures \cite{43,44,45}. One-shot NAS methods \cite{46,47,48} further reduced the cost by training large supernetworks and identifying high-accuracy subnetworks, often generated from pre-trained models. Nevertheless, as search spaces expand with architectural innovations \cite{49,10}, more efficient methods are necessary to predict neural network accuracy in vast design spaces.

Recent mathematical programming (MP) based NAS methods \cite{14,15} are noteworthy, as they transform multi-objective NAS problems into mathematical programming solutions. MP-NAS reduces search complexity from exponential to polynomial, presenting a promising avenue for large model architecture search. However, existing MP-NAS methods face architectural limitations. For instance, DeepMAD \cite{14} is designed for convolutional neural networks, while LayerNAS \cite{15} is suited for hierarchically connected networks. These limitations hinder MP-NAS usage in SOTA search spaces, leaving the challenge of swiftly designing effective large models unresolved. To address this, we propose an architecturally generalized MP-NAS, MathNAS. With the capability to perform rapid searches on any architecture, MathNAS presents an enticing approach to accelerate the design process for large models, providing a clearer and more effective solution.

\section{Theoretical Analysis}
\label{appendixA}

\subsection{$b_{i,j}^A$ Unbiased Proof}
The main text introduces the definition of $b_{i,j}^A$ as follows:
\begin{equation} 
	b^{A}_{i,j}=\overline{\Delta Acc\left(\mathcal{N}^{\Omega}_{(i,1) \rightarrow  (i,j)}\right )}= \cfrac{1}{n^{m-1}}\sum_{k=1}^{n^{m-1}}(Acc\left(\mathcal{N}^k_{(i,1)}\right)-Acc\left(\mathcal{N}^k_{(i,j)}\right))
\label{equ1}
\end{equation} 
In investigating the unbiasedness of $b_{i,j}^A$, it is essential to first examine the distributions of $ Acc\left(\mathcal{N}{(i,1)}\right)$ and $ Acc\left(\mathcal{N}{(i,j)}\right)$. We conducted an experiment on NAS-Bench-201 to discern the distribution of accuracies for all networks that include a specific block. Figure 1 illustrates the experimental results, which suggest that, when the Kolmogorov-Smirnov (KS) test is applied to a beta distribution, all p-values exceed 0.05. This allows us to infer that the network accuracy related to a specific block adheres to a beta distribution, with the parameters $\alpha$ and $\beta$ varying among different blocks.

We can thus hypothesize that $Acc\left(\mathcal{N}{(i,1)}\right)$ and $ Acc\left(\mathcal{N}{(i,j)}\right)$ follow the beta distributions with parameters $\alpha_{1}$, $\beta_{1}$ and $\alpha_{2}$, $\beta_{2}$, respectively. The expectation of $b_{i,j}^A$ is computed as follows:

\begin{figure}[h]
 \centering
 \begin{minipage}{0.23\linewidth}
     \centerline{\includegraphics[width=\textwidth]{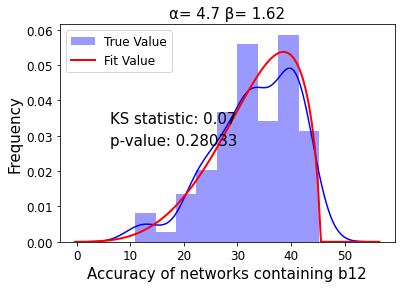}} 
 \end{minipage}
    \begin{minipage}{0.23\linewidth}
 
     \centerline{\includegraphics[width=\textwidth]{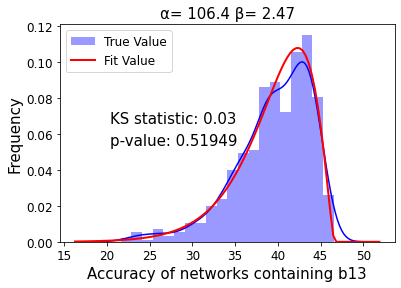}} 
 \end{minipage}
    \begin{minipage}{0.23\linewidth}
 
     \centerline{\includegraphics[width=\textwidth]{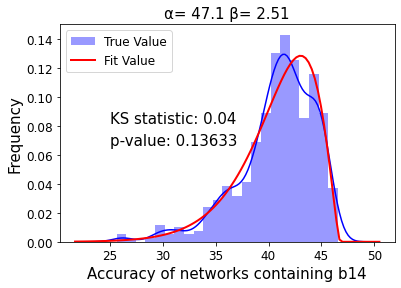}} 
   
 \end{minipage}
    \begin{minipage}{0.23\linewidth}
     \centerline{\includegraphics[width=\textwidth]{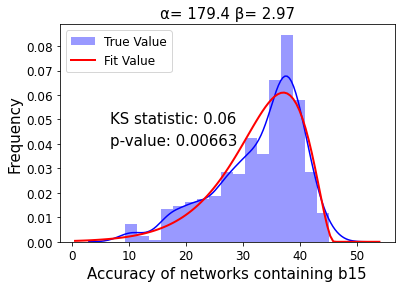}} 
 \end{minipage}
 
  \begin{minipage}{0.23\linewidth}
     \centerline{\includegraphics[width=\textwidth]{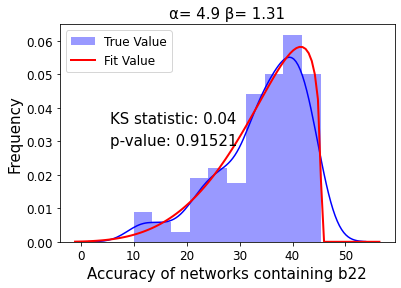}} 
 \end{minipage}
    \begin{minipage}{0.23\linewidth}
 
     \centerline{\includegraphics[width=\textwidth]{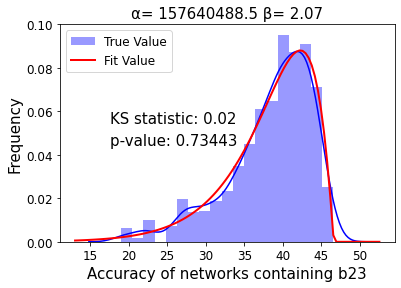}} 
 \end{minipage}
    \begin{minipage}{0.23\linewidth}
 
     \centerline{\includegraphics[width=\textwidth]{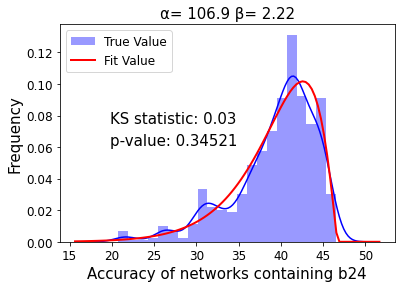}} 
   
 \end{minipage}
    \begin{minipage}{0.23\linewidth}
     \centerline{\includegraphics[width=\textwidth]{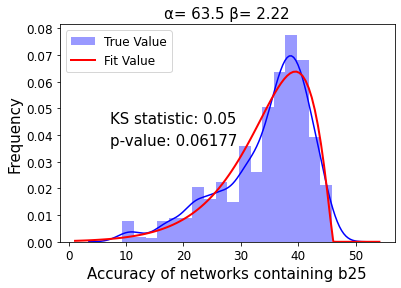}} 
 \end{minipage}

 \begin{minipage}{0.23\linewidth}
     \centerline{\includegraphics[width=\textwidth]{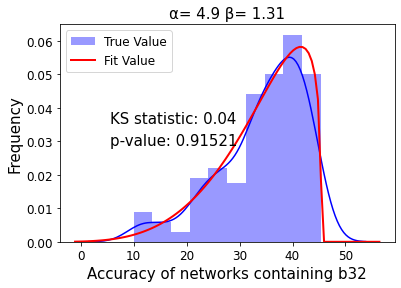}} 
 \end{minipage}
    \begin{minipage}{0.23\linewidth}
 
     \centerline{\includegraphics[width=\textwidth]{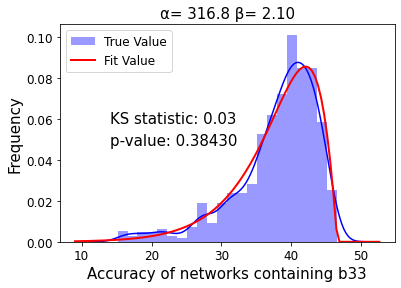}} 
 \end{minipage}
    \begin{minipage}{0.23\linewidth}
 
     \centerline{\includegraphics[width=\textwidth]{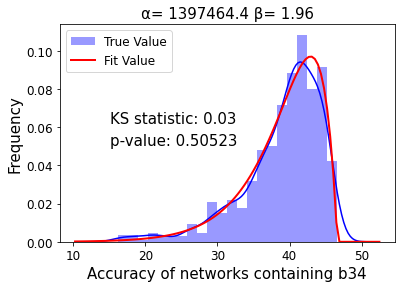}} 
   
 \end{minipage}
    \begin{minipage}{0.23\linewidth}
     \centerline{\includegraphics[width=\textwidth]{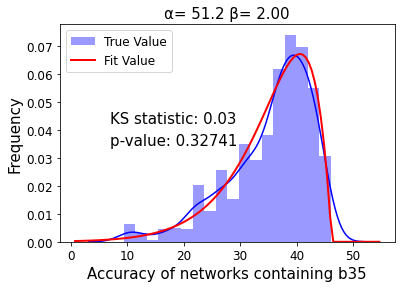}} 
 \end{minipage}
 \begin{minipage}{0.23\linewidth}
     \centerline{\includegraphics[width=\textwidth]{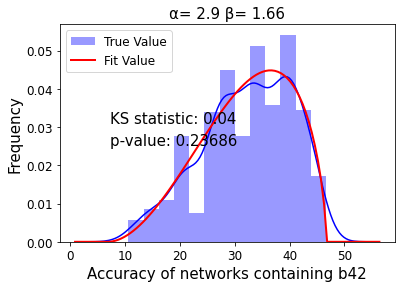}} 
 \end{minipage}
    \begin{minipage}{0.23\linewidth}
 
     \centerline{\includegraphics[width=\textwidth]{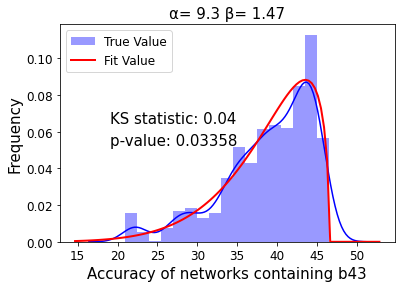}} 
 \end{minipage}
    \begin{minipage}{0.23\linewidth}
 
     \centerline{\includegraphics[width=\textwidth]{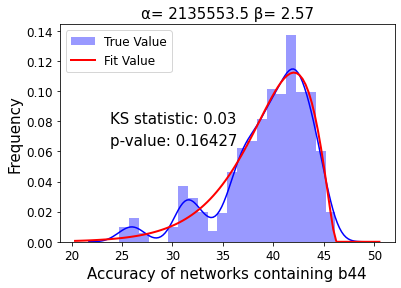}} 
   
 \end{minipage}
    \begin{minipage}{0.23\linewidth}
     \centerline{\includegraphics[width=\textwidth]{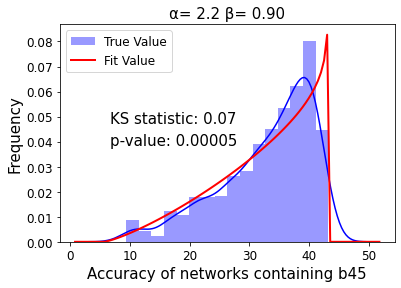}} 
 \end{minipage}
 \begin{minipage}{0.23\linewidth}
     \centerline{\includegraphics[width=\textwidth]{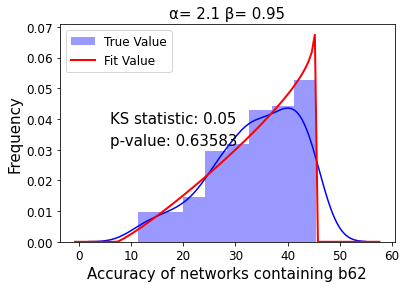}} 
 \end{minipage}
    \begin{minipage}{0.23\linewidth}
 
     \centerline{\includegraphics[width=\textwidth]{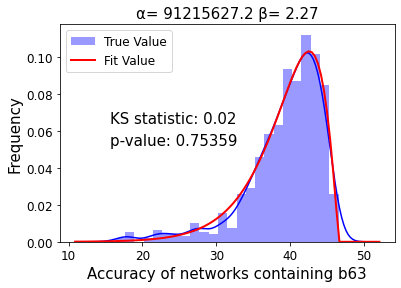}} 
 \end{minipage}
    \begin{minipage}{0.23\linewidth}
 
     \centerline{\includegraphics[width=\textwidth]{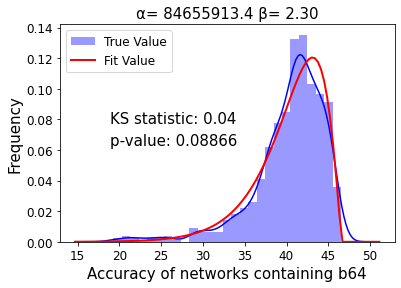}} 
   
 \end{minipage}
    \begin{minipage}{0.23\linewidth}
     \centerline{\includegraphics[width=\textwidth]{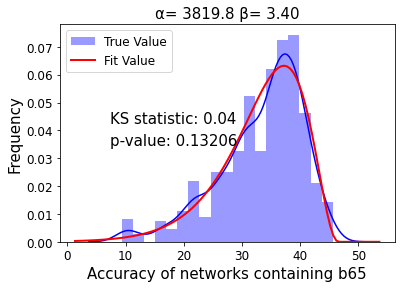}} 
 \end{minipage}
	\caption{ 
	Distribution of accuracy for networks containing specific blocks on NAS-Bench-201. Accuracy is the result of the network training on ImageNet for 200 epochs. It can be seen that the ks verification $p$ value of the beta distribution on all blocks exceeds 0.05, so it can be considered that the accuracy of all networks containing a specific block is in line with the beta distribution.
	}
	\label{unbisedproof}
\end{figure}

\begin{equation} 
\begin{split}
    E[b_{i,j}^A] &= E[\frac{1}{n^{m-1}} \sum_{k=1}^{n^{m-1}} (Acc\left(\mathcal{N}^k_{(i,1)}\right)-Acc\left(\mathcal{N}^k_{(i,j)}\right))] \\
    &=\frac{1}{n^{m-1}} \sum_{k=1}^{n^{m-1}} (E[Acc\left(\mathcal{N}^k_{(i,1)}\right)] - E[Acc\left(\mathcal{N}^k_{(i,j)}\right)])\\
    &=E[Acc\left(\mathcal{N}_{(i,1)}\right)] - E[Acc\left(\mathcal{N}_{(i,j)}\right)]\\
    &=\frac{\alpha_1}{\alpha_1+\beta_1} -\frac{\alpha_2}{\alpha_2+\beta_2}
\end{split} 
\label{equ2}
\end{equation} 
It is evident that the expected value of statistic $b_{i,j}^A$ represents the discrepancy between the expected values of $Acc\left(\mathcal{N}{(i,1)}\right)$ and $ Acc\left(\mathcal{N}{(i,j)}\right)$, and is independent of any specific sample value. In other words, regardless of the sample values observed, the expected value of statistic $b_{i,j}^A$ always reflects the difference between the expected values of $Acc\left(\mathcal{N}{(i,1)}\right)$ and $ Acc\left(\mathcal{N}{(i,j)}\right)$. Thus, we can assert that the statistic $b_{i,j}^A$ is unbiased, meaning the estimate of the difference between the expected values of $Acc\left(\mathcal{N}{(i,1)}\right)$ and $ Acc\left(\mathcal{N}{(i,j)}\right)$ remains unskewed.

Similarly, it can be argued that $b_{i,j}^L$ and $b_{i,j}^E$ are also unbiased. This conclusion adds to our understanding of the general applicability of these statistics in analyzing network accuracy associated with specific blocks.

\subsection{Fit Validation}

\begin{figure}[h]
 \centering
 \begin{minipage}{0.16\linewidth}
  
     \centerline{\includegraphics[width=\textwidth]{fig/2-1.png}} 
 \end{minipage}
    \begin{minipage}{0.16\linewidth}
 
     \centerline{\includegraphics[width=\textwidth]{fig/2-2.png}} 
 \end{minipage}
    \begin{minipage}{0.16\linewidth}
 
     \centerline{\includegraphics[width=\textwidth]{fig/2-3.png}} 
   
 \end{minipage}
    \begin{minipage}{0.16\linewidth}
     \centerline{\includegraphics[width=\textwidth]{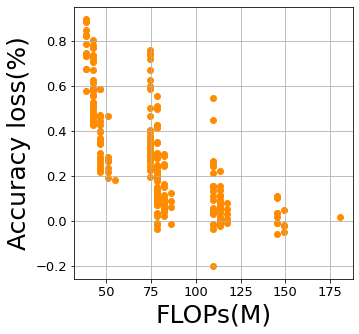}} 
 \end{minipage}
    \begin{minipage}{0.16\linewidth}
     \centerline{\includegraphics[width=\textwidth]{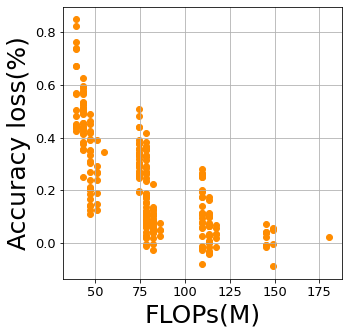}} 
 \end{minipage}
 \begin{minipage}{0.16\linewidth}
     \centerline{\includegraphics[width=\textwidth]{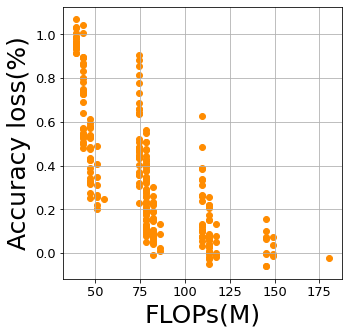}} 
 \end{minipage}
 \centerline{\small(a) $\Delta Acc(\mathcal{N}_{(i,1)\rightarrow (i,j)})$ vs $FLOPs(\mathcal{N})$ on NAS-Bench-201 (Cifar100).}
  \begin{minipage}{0.16\linewidth}
  
  
     \centerline{\includegraphics[width=\textwidth]{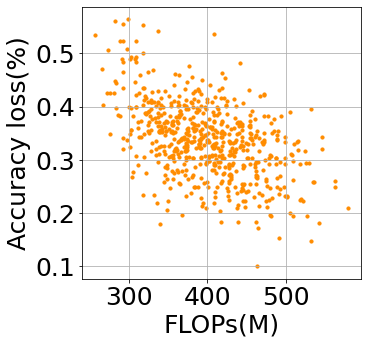}} 
 \end{minipage}
    \begin{minipage}{0.16\linewidth}
 
     \centerline{\includegraphics[width=\textwidth]{fig/3-2.png}} 
 \end{minipage}
    \begin{minipage}{0.16\linewidth}
 
     \centerline{\includegraphics[width=\textwidth]{fig/3-3.png}} 
   
 \end{minipage}
    \begin{minipage}{0.16\linewidth}
     \centerline{\includegraphics[width=\textwidth]{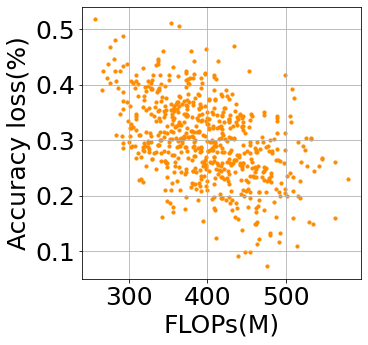}} 
 \end{minipage}
    \begin{minipage}{0.16\linewidth}
     \centerline{\includegraphics[width=\textwidth]{fig/3-5.png}} 
 \end{minipage}
     \begin{minipage}{0.16\linewidth}
     \centerline{\includegraphics[width=\textwidth]{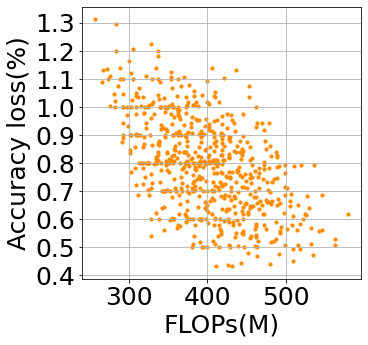}} 
 \end{minipage}
 \centerline{\small(b) $\Delta Acc(\mathcal{N}_{(i,1)\rightarrow (i,j)})$ vs $FLOPs(\mathcal{N})$ on MobileNetV3 (ImageNet).}
   \begin{minipage}{0.16\linewidth}
  
  
     \centerline{\includegraphics[width=\textwidth]{fig/4-1.png}} 
 \end{minipage}
    \begin{minipage}{0.16\linewidth}
 
     \centerline{\includegraphics[width=\textwidth]{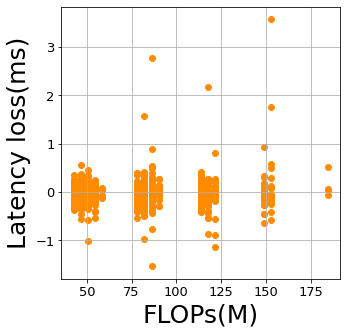}} 
 \end{minipage}
    \begin{minipage}{0.16\linewidth}
 
     \centerline{\includegraphics[width=\textwidth]{fig/4-3.png}} 
   
 \end{minipage}
    \begin{minipage}{0.16\linewidth}
     \centerline{\includegraphics[width=\textwidth]{fig/4-4.png}} 
 \end{minipage}
    \begin{minipage}{0.16\linewidth}
     \centerline{\includegraphics[width=\textwidth]{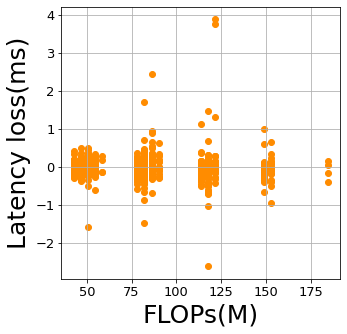}} 
 \end{minipage}
     \begin{minipage}{0.16\linewidth}
     \centerline{\includegraphics[width=\textwidth]{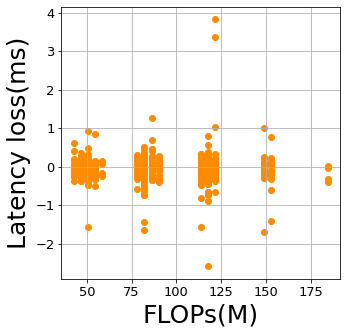}} 
 \end{minipage}
 \centerline{\small(c) $\Delta Lat(\mathcal{N}_{(i,1)\rightarrow (i,j)})$ vs $FLOPs(\mathcal{N})$ on NAS-Bench-201 (edgegpu).}
   \begin{minipage}{0.16\linewidth}
  
  
     \centerline{\includegraphics[width=\textwidth]{fig/5-1.png}} 
 \end{minipage}
    \begin{minipage}{0.16\linewidth}
 
     \centerline{\includegraphics[width=\textwidth]{fig/5-2.png}} 
 \end{minipage}
    \begin{minipage}{0.16\linewidth}
 
     \centerline{\includegraphics[width=\textwidth]{fig/5-3.png}} 
   
 \end{minipage}
    \begin{minipage}{0.16\linewidth}
     \centerline{\includegraphics[width=\textwidth]{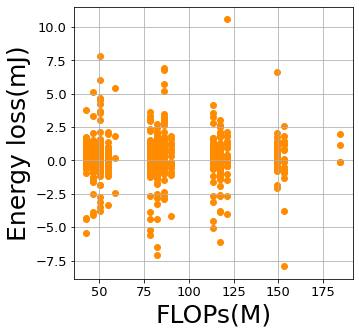}} 
 \end{minipage}
    \begin{minipage}{0.16\linewidth}
     \centerline{\includegraphics[width=\textwidth]{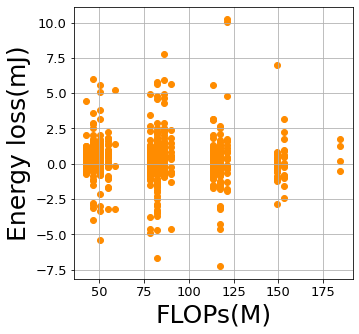}} 
 \end{minipage}
     \begin{minipage}{0.16\linewidth}
     \centerline{\includegraphics[width=\textwidth]{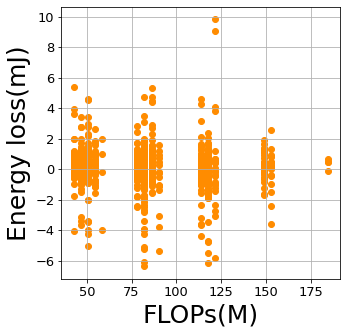}} 
 \end{minipage}
 \centerline{\small(d) $\Delta Eng(\mathcal{N}_{(i,1)\rightarrow (i,j)})$ vs $FLOPs(\mathcal{N})$ on NAS-Bench-201 (edgegpu).}
	\caption{ 
	More result of the relationship between $\Delta Acc(\mathcal{N}_{(i,1)\rightarrow (i,j)})$, $\Delta Lat(\mathcal{N}_{(i,1)\rightarrow (i,j)})$, $\Delta Eng (\mathcal{N}_{(i,1)\rightarrow (i,j)})$ and $FLOPs(\mathcal{N})$ .
	}
	\label{AppendixAmoreresult}
\end{figure}

Within the main body of the text, we put forth several rules derived from our empirical observations:
\begin{enumerate}
    \item For accuracy, $\Delta Acc(\mathcal{N}_{(i,1)\rightarrow (i,j_i)})\approx \alpha*1/\mathcal{F}(\mathcal{N})$.\\
    \item For latency and energy consumption, $\Delta Lat(\mathcal{N}{(i,1)\rightarrow (i,j_i)})$ and $\Delta Eng(\mathcal{N}{(i,1)\rightarrow (i,j_i)})$ remain relatively constant across different networks when referring to a specific block.
\end{enumerate}
We offer additional $\Delta Acc/Lat/Eng(\mathcal{N}{(i,1)\rightarrow (i,j)})$ - $FLOPs(\mathcal{N})$ pairs from NAS-Bench-201 and MobileNetV3 as depicted in Figure \ref{AppendixAmoreresult}, which serve to substantiate the universality of these rules. Furthermore, we fit $\Delta Acc(\mathcal{N}{(i,1)\rightarrow (i,j)})$ - $FLOPs(\mathcal{N})$ using different inverse functions. As evidenced in Table \ref{AppendixAfit}, a reciprocal function provides the best fit for this relationship.

\begin{table}[h]
\caption{Comparison of the fitting effect of different functions on the relationship between $\mathcal{N}_{(i,1)\rightarrow (i,j)}$ and $FLOPs(\mathcal{N})$.}
	\centering
    \resizebox{\linewidth}{!}{
	\begin{tabular}{c|ccc|ccc|ccc|ccc|ccc}
		\toprule 
		Net&\multicolumn{3}{c|}{Linear Function}&\multicolumn{3}{c|}{Quadratic Function}&\multicolumn{3}{c|}{Reciprocal Function}&\multicolumn{3}{c|}{Log Function}&\multicolumn{3}{c}{Exp Function}\\ 
		\midrule 
		~&$R^2$&MSE&DC&$R^2$&MSE&DC&$R^2$&MSE&DC&$R^2$&MSE&DC&$R^2$&MSE&DC\\
		\midrule
		MobileNetV3&0.47&0.0061&0.47&0.45&0.0062&0.46&0.51&0.0059&0.48&0.48&0.0061&0.48&0.001&0.011&0.0011\\
		\midrule
		NAS-Bench-201&0.57&0.018&0.57&0.46&0.022&0.46&0.68&0.013&0.68&0.61&0.015&0.61&0.004&0.04&0.004\\
		\bottomrule 
	\end{tabular}}
	\label{AppendixAfit}
\end{table}

\subsection{Ablation Analysis}
To elucidate the significance of each element within our predictive methodology, we instituted three sets of control models:
\begin{enumerate}
    \item An accuracy prediction model that operates without FLOPs information.
    \item A model predicting accuracy/latency where the Baseline network is arbitrarily chosen, instead of selecting a network with average FLOPs.
    \item A model predicting accuracy/latency that uses any arbitrary block as the baseline block as opposed to the first block.
\end{enumerate} 
Table \ref{appendixAablation} illustrates the outcomes of these models. It becomes evident that the omission of any rule adversely affects the model's performance. The second model, in particular, proves to be the most vital. These findings corroborate the necessity for the basenet to be the network of average FLOPs, aligning with the rule between $\Delta Acc(\mathcal{N}_{(i,1)\rightarrow (i,j)})$ -$FLOPs(\mathcal{N})$ we derived earlier.

\begin{table}[h] 
\centering
\caption{Ablation Study of the proposed prediction method on MobileNetV3}
\resizebox{0.7\linewidth}{!}{

\begin{tabular}{c|c|c|c|c}
		\toprule 
		\makecell{Predict\\Properties}&Network&RMSE&MAE&Error\\ 
		\midrule 
		\midrule
		\multirow{4}{*}{Accuracy}&Complete Model&0.117&0.0916&-\\ 
		\cdashline{2-5}[2pt/2pt]
		 \rule{0pt}{10pt}
		 	&Without FLOPs info&0.212&0.184&0.0924\\
		\cdashline{2-5}[2pt/2pt]
		\rule{0pt}{10pt}
		&Without Ave-FLOPs BaseNet&0.513&0.416&0.324\\
		\cdashline{2-5}[2pt/2pt]
		\rule{0pt}{10pt}
		&Without Super Baseblock&0.123&0.0992&0.0076\\
		\midrule
		\multirow{3}{*}{Latency}&Complete Model&0.268&0.189&-\\ 
		\cdashline{2-5}[2pt/2pt]
		\rule{0pt}{10pt}
		&Without Ave-FLOPs BaseNet&0.408&0.317&0.128\\
		\cdashline{2-5}[2pt/2pt]
		\rule{0pt}{10pt}
		&Without Super Baseblock&0.276&0.194&0.005\\
		\bottomrule 
	\end{tabular}}
   
	\label{appendixAablation}
\end{table}

\subsection{Effectiveness Analysis}
\begin{figure}[h]
  
    \centering 
	\begin{minipage}{0.3\linewidth}
		\hspace{5pt}
		\centerline{\includegraphics[width=\textwidth]{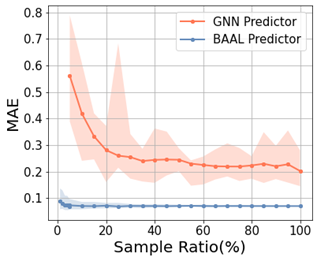}}
		 \centerline{\small(a) MobileNetV3}
	\end{minipage}
    \begin{minipage}{0.3\linewidth}
		\hspace{5pt}
		\centerline{\includegraphics[width=\textwidth]{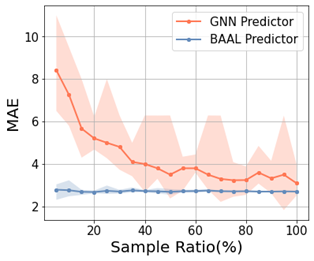}}
		\centerline{\small(b) NAS-Bench-201}
	\end{minipage}
    \caption{Sample effect of our accuracy predictor compared with GNN predictor.}
    \label{GNNcompare}
\end{figure}

This section provides an exploration into the efficacy of our proposed predictive model, substantiated by experimental validations. We draw comparisons with the classic GNN predictor on NAS-Bench-201 and MobileNetV3 search spaces. The GNN predictor's training process adheres to the Brp-NAS method. Figure \ref{GNNcompare} demonstrates that our model's MAE and MSE converge as the sample size of the network increases. This convergence is apparent in both weight-shared and weight-independent networks.

\subsection{Space Generality Analysis}
In our primary discussion, we reformulated the NAS problem as an Integer Linear Programming (ILP) problem. In this section, we delve further into the application of this formulation within the actual search space.

\paragraph{Flexibility in Number of Block Nodes.} Some search spaces provide the flexibility to choose the number of block nodes. For instance, the SuperTransformer allows for the selection of anywhere between 1 and 6 decoders. For this kind of search space, MathNAS can also be applied. Specifically, for block node $i$ that has no block operation, we regard it as a special block of this node and represent it with $b_{i,0}$. Under this condition, the ILP formula can be modified accordingly as follows:
\begin{equation}
\begin{split}
     &\max \limits_{b^B}Acc(\widetilde{\mathcal{N}}) - \cfrac{\sum_{i=1}^m\sum_{j=0}^nb_{i,j}^A*b_{i,j}^B}{\sum_{i=1}^m\sum_{j=1}^nb_{i,j}^F*b_{i,j}^B}*FLOPs(\overline{\mathcal{N}})\\
     &s.t.\\
    &Lat(\widetilde{\mathcal{N}})-\sum_{i=1}^m\sum_{j=0}^nb_{i,j}^L*b_{i,j}^B\leq \hat{L},\\
    &Eng(\widetilde{\mathcal{N}})-\sum_{i=1}^m\sum_{j=0}^nb_{i,j}^E*b_{i,j}^B\leq \hat{E},\\
    &\sum_{k=1}^i b_{k,0}^B+\sum_{j=1}^nb_{i,j}^B=1, b_{i,j}^B\in\left\{0,1\right\}, \forall{1\leq i \leq m}.
\end{split}
\label{eq14}
\end{equation}

\paragraph{Applicability in Micro Search Space.}
Unlike other search spaces we use, NAS-Bench-201 is a micro search space. 
In this paragraph, we explain how we performed our experiments in this search space and why MathNAS is applicable.

\textit{Practical Implementation (How)}: During experiments within the NAS-Bench-201 space, we identify a set of edges in the same position across multiple GNN cells as a single "block". Given that the structure of GNN cells in the network remains consistent, our focus is on a single cell. Hence, any alteration in an edge operation essentially translates to a corresponding change in all cells, while other edge operations remain static.

\textit{Theoretical Framework (Why)}: Regardless of the distinction between macro and micro search spaces, the networks in both are assembled from multiple mutable modules, be it blocks or edges. The capability of the entire network can be represented by the capabilities of these individual modules. In MathNAS, to explore the contribution of module capabilities to the network's performance, we evaluated changes in inherent module capabilities and their interactive capacities during module switches. This module evaluation methodology is applicable to the definition of blocks in NAS-Bench-201 that we proposed above: alterations in edge operations impact not only the specific edges' output data (inherent capability) but also influence the input and output data of other edges within the network (interactive capability). Therefore, MathNAS is theoretically suitable for micro search spaces represented by NAS-Bench-201.

\subsection{Searching Equation Solving Details}
In this section, we describe in detail the solution of the fractional objective function programming equation proposed in the paper. 
The solution is divided into two steps. 
\begin{enumerate}
    \item Convert the original equation into an integer linear programming equation.
    \item Solve the ILP equation.
\end{enumerate}
\par \bigskip
\textbf{Equation Transformation.}
In order to transform the equation into an ILP problem, we first perform variable substitution on the original equation.
\begin{equation} 
     let\quad b^{\widetilde{B}}_{i,j}=\cfrac{b^B_{i,j}}{\sum_{i=1}^m\sum_{i=1}^nb^F_{i,j}*b^B_{i,j}},\quad z=\cfrac{1}{\sum_{i=1}^m\sum_{i=1}^nb^F_{i,j}*b^B_{i,j}}
\end{equation}
Then the original equation can be transformed into the following integer linear programming problem:
\begin{equation}
\begin{split}
     &O = \min\limits_{b^{\widetilde{B}},z}{ (\sum_{i=1}^m\sum_{j=1}^nb^A_{i,j}*b^{\widetilde{B}}_{i,j}*\overline{\mathcal{F}(\mathcal{N})})}\\
     &s.t.\\
     &(Lat(\widetilde{\mathcal{N}})-\hat{L})*z \leq \sum_{i=1}^m\sum_{j=1}^nb^L_{i,j}*b^{\widetilde{B}}_{i,j},  (Eng(\widetilde{\mathcal{N}})-\hat{E})*z \leq \sum_{i=1}^m\sum_{j=1}^nb^E_{i,j}*b^{\widetilde{B}}_{i,j} \\
     &\forall{1\leq i \leq m}, \sum_{j=1}^nb^{\widetilde{B}}_{i,j}=z, b^{\widetilde{B}}_{i,j}\in\left\{0,z\right\}.
\end{split} 
\label{equ16}
\end{equation}
\par \bigskip
\textbf{ILP Solving.}
To solve the ILP equations, we use the off-the-shelf Linprog Python package and the Gurobipy Python package to find feasible candidate solutions.
\begin{itemize}
    \item Linprog is a basic integer programming solver that can be used on almost all edge devices, even on the resource-constrained Raspberry Pi. We use it to implement the branch and bound method and solve the ILP problem.
    \item Gurobipy is a more powerful solver, which has built-in a variety of advanced solving algorithms such as heuristic algorithms, and can flexibly utilize all available hardware resources on the device. Although Gurobipy is powerful, it requires more hardware resources than Linprog.
\end{itemize}
Therefore, for devices with limited hardware resources, we use Linprog for searching. For well-resourced devices, we use Gurobipy.

\newpage

\section{Experimental Setups}
\label{appendixB}
\subsection{Description of Search Space}
This section provides a detailed overview of the search spaces that we employ in our experimental setup.
Furthermore, in our experiments, each resolution corresponds to a separate search space.

\textbf{NAS-Bench-201} search space encompasses cell-based neural architectures, where an architecture is represented as a graph. Each cell in this graph comprises four nodes and six edges, with each edge offering a choice among five operational candidates - zerorize, skip connection, 1-by-1 convolution, 3-by-3 convolution, and 3-by-3 average pooling. This design leads to a total of 15,626 unique architectures. The macro skeleton is constructed with one stem cell, three stages each composed of five repeated cells, residual blocks between the stages, and a final classification layer which incorporates an average pooling layer and a fully connected layer with a softmax function. The stem cell is formed by a 3-by-3 convolution with 16 output channels followed by a batch normalization layer. Each cell within the three stages has 16, 32 and 64 output channels respectively. The intermediate residual blocks contain convolution layers with stride 2 for down-sampling.

\textbf{MobileNetV3} search space follows a layer-by-layer paradigm, where the building blocks use MBConvs, squeeze and excite mechanisms, and modified swish nonlinearity to construct efficient neural networks. This space is organized into five stages, each containing a number of building blocks varying from 2 to 4. The kernel size for each block can be chosen from \{3, 5, 7\} and the expansion ratio from \{1, 4, 6\}. The search space encapsulates approximately $10^{19}$ sub-nets, with each block offering 7,371 choices.

\textbf{SuperViT} search space is a tiled space featuring a fixed macroarchitecture. It is composed of three dynamic CNN blocks and four dynamic Transformer blocks connected in sequence. Each block allows for variations in width, depth, kernel size and expansion ratio. The SuperViT search space is displayed in Table \ref{nasvitspace}.

\begin{table}[h]
\caption{An illustration of SuperViT search space. Bold black represents blocks selected for the baseline net. The block in basenet corresponds to the largest of all available options.}
	\centering
	\resizebox{1\linewidth}{!}{
	\begin{tabular}{c|ccccccc}
		\toprule 
		Block&Width&Depth&Kernel size&Expansion ratio&SE&Stride&Number of Windows\\ 
        \midrule
        Conv&$\left\{\textbf{16},24\right\}$&-&3&-&-&2&-\\
        MBConv-1&$\left\{16,\textbf{24}\right\}$&$\left\{\textbf{1},2\right\}$&$\left\{\textbf{3},5\right\}$&1&N&1&-\\
        MBConv-2&$\left\{24,\textbf{32}\right\}$&$\left\{\textbf{3},4,5\right\}$&$\left\{\textbf{3},5\right\}$&$\left\{4,5,\textbf{6}\right\}$&N&2&-\\
        MBConv-3&$\left\{\textbf{32},40\right\}$&$\left\{3,4,5,\textbf{6}\right\}$&$\left\{\textbf{3},5\right\}$&$\left\{4,\textbf{5},6\right\}$&Y&2&-\\
        Transformer-4&$\left\{\textbf{64},72\right\}$&$\left\{3,4,\textbf{5},6\right\}$&-&$\left\{\textbf{1},2\right\}$&-&2&1\\
        Transformer-5&$\left\{\textbf{112},120,128\right\}$&$\left\{3,4,5,\textbf{6},7,8\right\}$&-&$\left\{\textbf{1},2\right\}$&-&2&1\\
        Transformer-6&$\left\{160,168,176,\textbf{184}\right\}$&$\left\{3,4,5,\textbf{6},7,8\right\}$&-&$\left\{\textbf{1},2\right\}$&-&1&1\\
        Transformer-7&$\left\{\textbf{208},216,224\right\}$&$\left\{3,4,\textbf{5},6\right\}$&-&$\left\{1,\textbf{2}\right\}$&-&2&1\\
        MBPool&$\left\{1792,\textbf{1984}\right\}$&-&1&6&-&-&-\\
        \midrule
        Input Resolution&\multicolumn{7}{c}{$\left\{192,224,256,288\right\}$}\\
		\bottomrule 
	\end{tabular}}
	\label{nasvitspace}
\end{table}

\textbf{SuperTransformer} search space we employ is a large design space, constructed with Arbitrary Encoder-Decoder Attention and Heterogeneous Layers. This space contains an encoder layer and a decoder layer, the embedding dimensions of which can be selected from {512, 640}. The encoder comprises six layers, each with an attention module and an FFN layer in series. The number of decoder layers can range from 1 to 6, and each decoder layer consists of two attention modules and one FFN layer in series. Each decoder has the option to focus on the last, second last or third last encoder layer. The hidden dimension of the FFN can be chosen from {1024, 2048, 3072}, and the number of heads in attention modules from {4, 8}. Every encoder layer has six choices, while each decoder layer provides 36 options. The search space and its corresponding baseline network are illustrated in Table \ref{transferspace}.

\begin{table}[h]
\caption{An illustration of SuperTransformer search space and the baseline net.}
	\centering
	\resizebox{0.6\linewidth}{!}{
	\begin{tabular}{c|c|c}
		\toprule 
		Module&Choice&Baseline Net\\
        \midrule
        encoder-embed&[640, 512]&512\\
        decoder-embed&[640, 512]&512\\

encoder-ffn-embed-dim& [3072, 2048, 1024]&2048\\
decoder-ffn-embed-dim&[3072, 2048, 1024]&2048\\

encoder-layer-num & [6]&6\\
decoder-layer-num & [6, 5, 4, 3, 2, 1]&6\\

encoder-self-attention-heads & [8, 4]&4\\
decoder-self-attention-heads & [8, 4]&4\\
decoder-ende-attention-heads & [8, 4]&4\\
 
decoder-arbitrary-ende-attn & [-1, 1, 2]&1\\
 
		\bottomrule 
	\end{tabular}}
	\label{transferspace}
\end{table}

\section{Additional Experiments}
\label{appendixC}
\subsection{Experiment on NAS-Bench-201 Search Space}

In the main body of this paper, we have showcased the competency of MathNAS in attaining accuracy-latency trade-offs within the NAS-Bench-201 search space. In this section, we extend our discussion to compare MathNAS with the current state-of-the-art models.

A comparative analysis with the leading-edge algorithms on NAS-Bench-201 is presented in Table \ref{nas-bench-201}. All the algorithms under consideration utilize the CIFAR-10 training and validation sets for architectural search and employ the NAS-bench-201 API to ascertain the ground-truth performance of the searched architecture across the three datasets. The reported experimental results are averaged over four separate searches.

From the results, it is evident that our method outperforms the state-of-the-art methods, with our best results approaching the peak performance. This robust performance serves as a testament to the efficacy of our proposed approach.

\begin{table}[h]
\caption{Comparison results of MathNAS with state-of-the-art NAS methods on NAS-Bench-201.}
	\centering
	\resizebox{1\linewidth}{!}{
	\begin{tabular}{ccccccc}
		\toprule 
		\multirow{2}{*}{Method}&\multicolumn{2}{c}{CIFAR-10}&\multicolumn{2}{c}{CIFAR-100}&\multicolumn{2}{c}{ImageNet-16-120}\\ 
	    \cline{2-7}
	    \rule{0pt}{10pt}
		&validation&test&validation&test&validation&test\\
        \midrule
		$DARTS^{1st}$&$39.77\pm0.00$&$54.30\pm0.00$&$15.03\pm0.00$&$15.61\pm0.00$&$16.43\pm0.00$&$16.32\pm0.00$\\
		$DARTS^{2nd}$&$39.77\pm0.00$&$54.30\pm0.00$&$38.57\pm0.00$&$38.97\pm0.00$&$18.87\pm0.00$&$18.41\pm0.00$\\
		SETN&$84.04\pm0.28$&$87.64\pm0.00$&$58.86\pm0.06$&$59.05\pm0.24$&$33.06\pm0.02$&$32.52\pm0.21$\\ 
		FairNAS&$90.07\pm0.57$&$93.23\pm0.18$&$70.94\pm0.84$&$71.00\pm1.46$&$41.9\pm1.00$&$42.19\pm0.31$\\
		SGNAS&$90.18\pm0.31$&$93.53\pm0.12$&$70.28\pm1.20$&$70.31\pm1.09$&$44.65\pm2.32$&$44.98\pm2.10$\\
		DARTS-&$91.03\pm0.44$&$93.80\pm0.40$&$71.36\pm1.51$&$71.53\pm1.51$&$44.87\pm1.46$&$45.12\pm0.82$\\
		\midrule
		\textbf{Ours}&\textbf{90.18}$\pm$\textbf{0.00}&\textbf{93.31}$\pm$\textbf{0.00}&\textbf{71.74}$\pm$\textbf{0.00}&\textbf{70.82}$\pm$\textbf{0.00}&\textbf{46.00}$\pm$\textbf{0.00}&\textbf{46.53}$\pm$\textbf{0.00}\\
		Optimal&$91.61$&$94.37$&$73.49$&$73.51$&$46.77$&$47.31$\\
		\bottomrule 
	\end{tabular}}
	\label{nas-bench-201}
\end{table}

\subsection{Performance Comparison with Block-wise Methods}
Existing blockwise methods such as DNA \cite{57}, DONNA \cite{58}, and LANA \cite{59} use block distillation techniques to block the teacher model, obtaining the architecture of blocks to be replaced and their performance evaluation. Following this, they use the performance of each block to guide the algorithm in finding models with superior performance.
Recent work \cite{60} has pointed out the limitations of such methods. They depend on an excellent teacher network due to their use of distillation techniques. Furthermore, previous block performance estimation methods are unable to effectively predict actual block performance. Additionally, these methods are only suitable for the macro search space. 
In this section, we compare MathNAS with these methods and demonstrate that MathNAS overcomes these limitations.

We compare the performance of our method and previous blocking methods on the MobileNetV3 search space and the ImageNet dataset.
For DNA and DONNA, we use the data reported in the original DONNA paper, where DONNA selects the data of the predictor after training on 40 samples. In the case of LANA, due to the absence of the official code, we computed the metrics using the supernet in the search space as the teacher net, determining block evaluation scores from the variation in verification accuracy. The final network performance scores result from a linear addition, in line with LANA's methodology. 
The results are shown in Table \ref{appendixblockwosed}.
Evidently, MathNAS outperforms prior block-wise methods in both network performance evaluation and accuracy prediction.

\begin{table}[h] 
\centering
\caption{
Comparison of prediction accuracy between our proposed method and block-wise methods on the MobileNetV3 (x 1.2) search space with the ImageNet dataset.}

\resizebox{\linewidth}{!}{

\begin{tabular}{c|c|c|c|c}
		\toprule 
		Methods&Kendall-Tau&MSE&Block Evaluation&Net Evaluation\\ 
		\midrule 
		DNA&0.74&NA&Block Knowledge Distillation&Sorting Model\\ 
            DONNA&0.82&0.08&Distillation Loss&Linear Regression\\
            LANA&0.77&0.04&Change of Validation Accuracy&Simple Summation\\
            \cdashline{1-5}[2pt/2pt]
		\rule{0pt}{10pt}
            \textbf{MathNAS}&\textbf{0.86}&\textbf{0.01}&\textbf{Average of Accuracy Variations}&\textbf{MathNAS Formula}\\
		
		\bottomrule 
	\end{tabular}}
   
	\label{appendixblockwosed}
\end{table}

\begin{figure}[h]
	\centering
	\begin{minipage}{0.32\linewidth}
		 
		\centerline{\includegraphics[width=\textwidth]{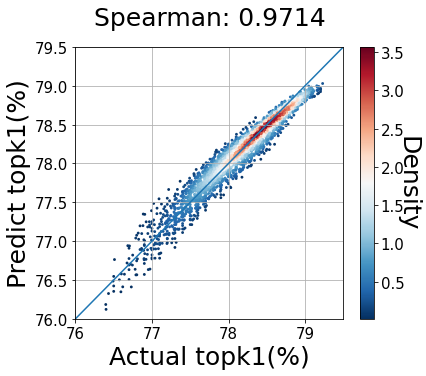}}
         \centerline{\small(a) MathNAS}
	\end{minipage}
	\begin{minipage}{0.32\linewidth}		
		\centerline{\includegraphics[width=\textwidth]{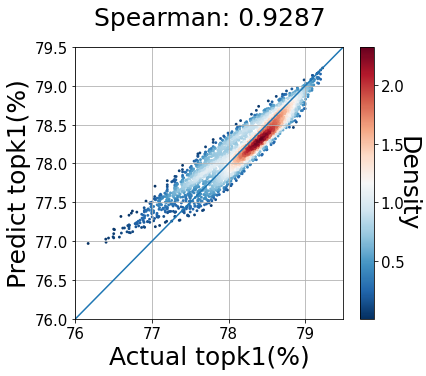}}
           \centerline{\small(b) LANA}
	\end{minipage}
        \begin{minipage}{0.31\linewidth}		
		\centerline{\includegraphics[width=\textwidth]{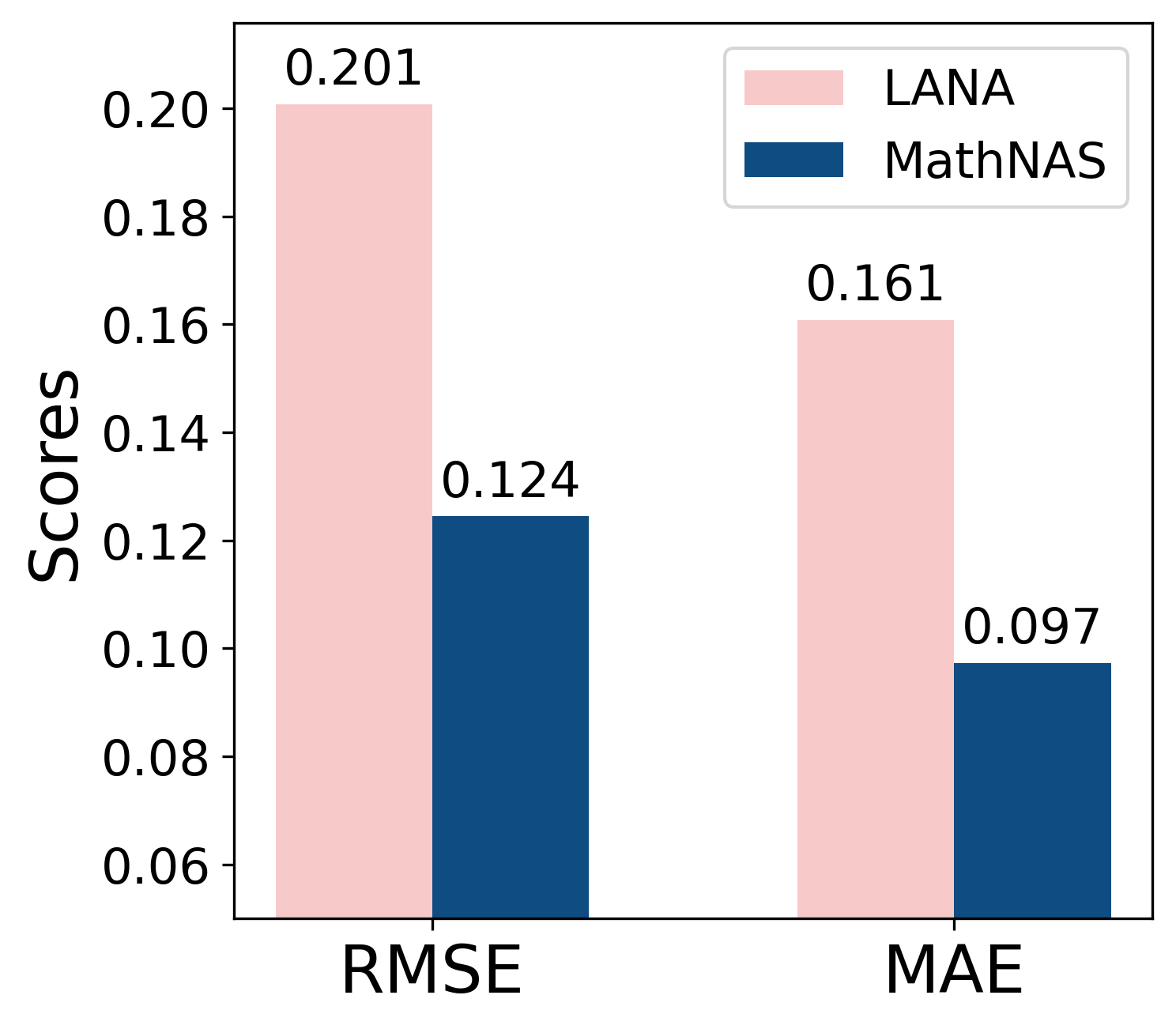}}
          \centerline{\small(c) Comparation}
	\end{minipage}
	\caption{Evaluation of accuracy prediction capabilities of MathNAS versus LANA in the MobileNetV3 (x 1.2) search space on ImageNet.}
	\label{appendixlalamobile}
\end{figure}

\begin{figure}[h]
	\centering
	\begin{minipage}{0.32\linewidth}
	 	\centerline{\includegraphics[width=\textwidth]{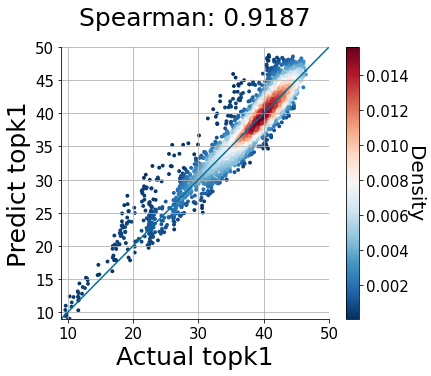}}
        \centerline{\small(a) MathNAS}
        
	\end{minipage}
	\begin{minipage}{0.32\linewidth}		
		
        \hspace{10pt}
        \centerline{\includegraphics[width=\textwidth]{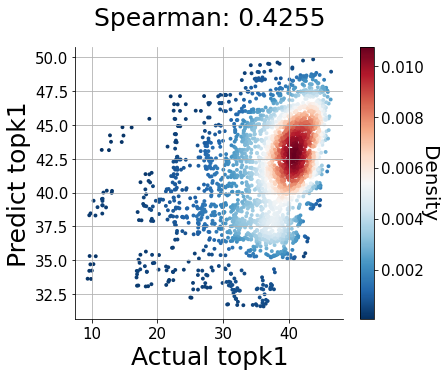}}
          \centerline{\small(b) LANA}
	\end{minipage}
        \hspace{5pt}
        \begin{minipage}{0.28\linewidth}		
		\centerline{\includegraphics[width=\textwidth]{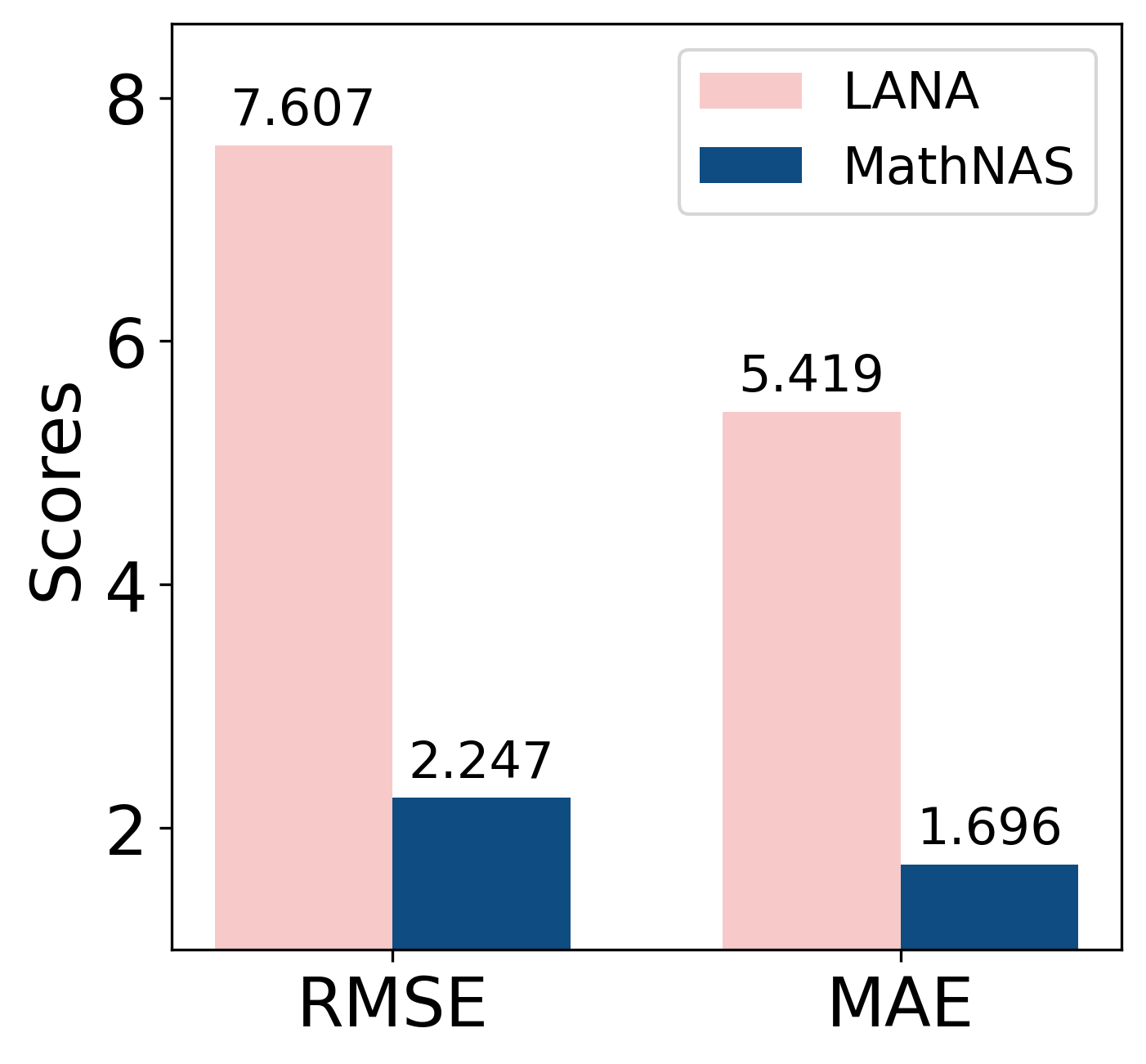}}
           \centerline{\small(c) Comparation}
	\end{minipage}
	\caption{Accuracy prediction comparison between MathNAS and LANA within the NAS-Bench-201 search space. }
	\label{appendixlalanasbench201}
\end{figure}

Previous blocking methods were targeted at macrospaces, while MathNAS has excellent performance in both macrospaces and microspaces. Figures \ref{appendixlalamobile} and \ref{appendixlalanasbench201} show the performance comparison between MathNAS and LANA on macrospace MobileNetV3 and microspace NAS-Bench-201 respectively.
Accuracy refers to ImageNet validation after 200 independent training epochs. For LANA's assessment, the largest FLOPs network in NAS-Bench-201 serves as the teacher net, with validation accuracy changes noted for each block. 
As shown in Figure \ref{appendixlalanasbench201}(b), while LANA is tailored for macro spaces and struggles in micro spaces, MathNAS exhibits consistent prediction accuracy even in the micro-space NAS-Bench-201, underscoring its search space adaptability.

Overall, MathNAS solves the following limitations compared to previous blockwise methods:
\begin{itemize}
    \item It does not depend on distillation but proposes a new block evaluation method based on the observed relationship between FLOPs and delta-Accuracy. This method is mathematically efficient and succinct, and it has been theoretically validated across different search spaces.
    \item It applies to a wider variety of search spaces, beyond the classical ones apart from the macro search space, such as the micro search space of NB201.
    \item Its evaluation of block performance and network accuracy prediction is more precise.
    \item It can find more superior architectures based on a full-space search, holding true even when compared to non-blockwise NAS methods. Moreover, the time complexity of the search algorithm is at a polynomial level.
\end{itemize}

\subsection{Experiment on Dynamic Network}
In the main body of this paper, we analyze the dynamic performance of MathNAS and state-of-the-art (SOTA) dynamic networks on real devices. Here, we extend this analysis by providing a more comprehensive comparison of the on-device performance of MathNAS and SOTA dynamic networks.

\begin{figure}[h]
	\centering
	\begin{minipage}{0.4\linewidth}
		\hspace{5pt}
		\centerline{\includegraphics[width=\textwidth]{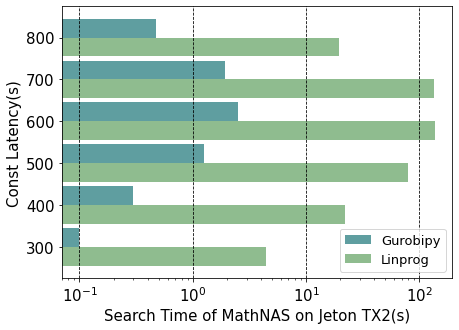}}
	  
	\end{minipage}
	\begin{minipage}{0.5\linewidth}
		
		\centerline{\includegraphics[width=\textwidth]{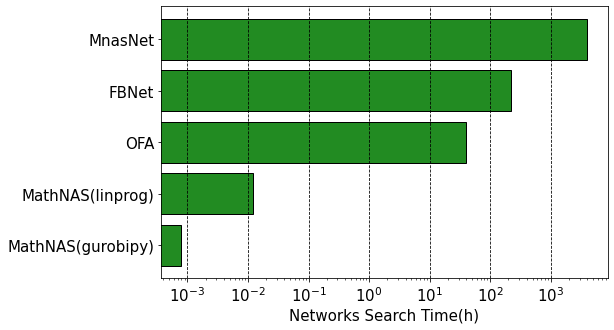}}
	\end{minipage}
	\caption{Experiments are performed on OFA search space and Jetson TX2 GPU. (a) Search time of MathNAS under different latency constraints. (b) Comparison of search times of MathNAS with other NAS methods.}
	\label{afig3}
\end{figure}

\paragraph{Search Time.} We delve into the search time required by MathNAS under various latency constraints, comparing it with other methodologies. Figure \ref{afig3} (a) reveals that the search time varies under different latency constraints. More specifically, when the latency constraint is either relatively large or small, the search duration decreases. This is primarily due to the pruning process undertaken using the branch and bound method, which improves search efficiency by employing upper and lower bounds. A larger latency constraint results in a higher lower bound, and a smaller latency constraint results in a lower upper bound. This in turn increases the extent of pruning, reducing the search range and thus enhancing the search speed. Moreover, we compare the search times of various NAS methods. As Figure \ref{afig3} (b) indicates, the search time on Linprog is longer than that on Gurobipy. This is because Gurobipy can utilize all available hardware resources on a device, whereas Linprog relies solely on a single CPU. Despite this, Linprog's search time is about 1/3300 of that of OFA, resulting in a significant reduction in NAS search time. For Gurobipy, the search time is 1/80000 of the OFA search time.
 
\begin{figure}[h]
 \centering
 \begin{minipage}{0.3\linewidth}
     \centerline{\includegraphics[width=\textwidth]{fig/6-1.png}}
 \end{minipage}
 \begin{minipage}{0.3\linewidth}
     \centerline{\includegraphics[width=\textwidth]{fig/6-2.png}}
 \end{minipage}
    \begin{minipage}{0.3\linewidth}
     \centerline{\includegraphics[width=\textwidth]{fig/6-3.png}}
 \end{minipage}
 \begin{minipage}{0.3\linewidth}
     \centerline{\includegraphics[width=\textwidth]{fig/6-4.png}}
 \end{minipage}
 \begin{minipage}{0.3\linewidth}
     \centerline{\includegraphics[width=\textwidth]{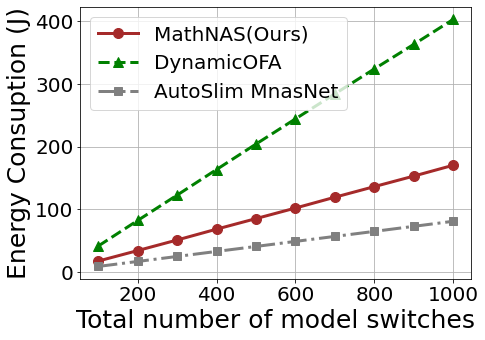}}
 \end{minipage}
 \begin{minipage}{0.3\linewidth}
     \centerline{\includegraphics[width=\textwidth]{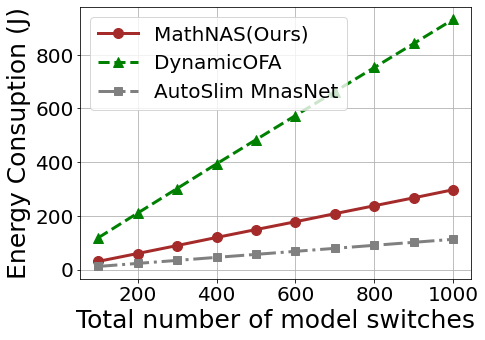}}
 \end{minipage}
     
	\caption{Latency (top) and energy consumption (bottom) required for different dynamic networks switching on Raspberry Pi (left), TX2GPU (middle) and TX2CPU (right).}
	\label{fig6}
\end{figure}

\textbf{Energy.} The key objective of dynamic networks is to adaptively switch architectures based on the device environment. As such, the time and energy consumed during architecture switching, as well as the ability to respond to environmental changes, are crucial metrics for dynamic network methods. In this section, we scrutinize the performance of MathNAS in dynamic environments with latency and energy as metrics. We also utilize the UNI-T UT658 power monitor to gauge energy consumption during the switching process. Initial findings, as displayed in Figure \ref{fig6}, indicate that both the switching energy and latency of MathNAS are significantly less than those of Dynamic-OFA. This is primarily due to the fact that MathNAS only switches necessary sub-blocks, while Dynamic-OFA switches entire sub-nets. When deployed on Raspberry Pi, MathNAS requires approximately 80\% less energy to switch sub-nets compared to Dynamic-OFA. Furthermore, the latency and energy required for MathNAS to switch sub-nets are only marginally higher than those required by AutoSlim, despite the fact that the size of candidate networks in AutoSlim is much smaller than in MathNAS.

\newpage
\section{Searched Architecture Visualization}
\label{appendixD} 

In this section, we provide visualizations of our searched architectures, including MathNAS-MB and MathNAS-T, and dynamic networks.

\subsection{Visualization Architecture on MobileNet-V3}
\begin{figure}[h]
 
 \begin{minipage}{0.6\linewidth}
     \centerline{\includegraphics[width=\textwidth]{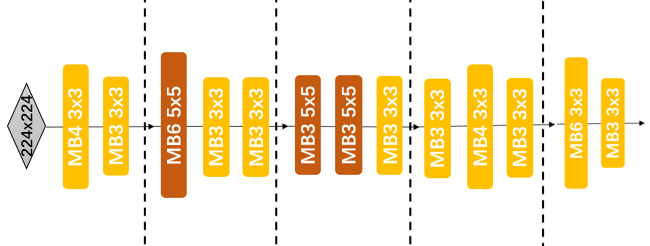}}
 \end{minipage}
 
 \centerline{\small(a) Architecture Visualization of MathNAS-MB1}
 
  \vspace{20pt}
 \begin{minipage}{0.5\linewidth}
     \centerline{\includegraphics[width=\textwidth]{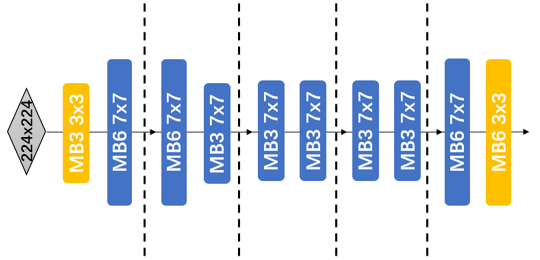}}
 \end{minipage}
 
 \centerline{\small(b) Architecture Visualization of MathNAS-MB2}
 
  \vspace{20pt}
\begin{minipage}{0.8\linewidth}
     \centerline{\includegraphics[width=\textwidth]{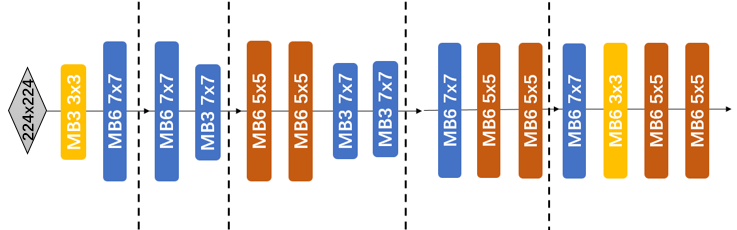}}
 \end{minipage}
 \centerline{\small(c) Architecture Visualization of MathNAS-MB3}
 
 \vspace{20pt}
 \begin{minipage}{1\linewidth}
     \centerline{\includegraphics[width=\textwidth]{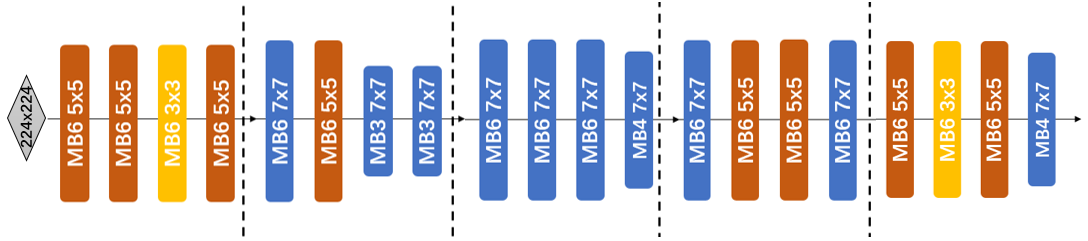}}
 \end{minipage}
 \centerline{\small(d) Architecture Visualization of MathNAS-MB4}
	\caption{Architecture Visualization of MathNAS-MB Network.}
	\label{MB}
\end{figure}

Figure \ref{MB} shows the visual architecture of the network searched by MathNAS on the MobileNet-V3 search space.

\subsection{Visualization Architecture on SuperViT Space}
Table \ref{nasvit} shows the visual architecture of the network searched by MathNAS on the SuperViT search space.

\begin{table}[h]
\caption{Architecture visualization of MathNAS-T model. ‘c’ denotes the number of output channels, ‘d’ denotes the number of layers, ‘ks’ denotes
kernel size, ‘e’ denotes expansion ratio, ‘k’ denotes the number of windows, and ‘s’ denotes stride.}
	\centering
	\resizebox{1\linewidth}{!}{
	\begin{tabular}{c|ccccc}
		\toprule 
	     &MathNAS-T1&MathNAS-T2&MathNAS-T3&MathNAS-T4&MathNAS-T5\\
	     \midrule
	     \multirow{4}{*}{Conv}&c:16&c:16&c:24&c:16&c:24\\
	     ~&d:1&d:1&d:1&d:1&d:1\\ 
	     ~&ks:3&ks:3&ks:3&ks:3&ks:3\\
	     ~&s:2&s:2&s:2&s:2&s:2\\
	     \midrule
	     \multirow{5}{*}{MBConv-1}&c:16&c:16&c:16&c:16&c:24\\
	     ~&d:3&d:3&d:1&d:1&d:2\\ 
	     ~&ks:3&ks:3&ks:&ks:5&ks:3\\
	     ~&e:1&e:1&e:1&e:1&e:1\\
	     ~&s:1&s:1&s:1&s:1&s:1\\
	     \midrule
	     \multirow{5}{*}{MBConv-2}&c:24&c:24&c:24&c:24&c:24\\
	     ~&d:3&d:4&d:3&d:3&d:4\\ 
	     ~&ks:3&ks:3&ks:&ks:3&ks:3\\
	     ~&e:4&e:4&e:5&e:5&e:5\\
	     ~&s:2&s:2&s:2&s:2&s:2\\
	     \midrule
	     \multirow{5}{*}{MBConv-3}&c:32&c:40&c:40&c:32&c:40\\
	     ~&d:3&d:3&d:4&d:6&d:6\\ 
	     ~&ks:3&ks:3&ks:&ks:3&ks:3\\
	     ~&e:4&e:4&e:4&e:4&e:5\\
	     ~&s:2&s:2&s:2&s:2&s:2\\
	     \midrule
	     \multirow{5}{*}{Transformer-4}&c:64&c:64&c:64&c:72&c:72\\
	     ~&d:3&d:4&d:3&d:4&d:6\\ 
	     ~&k:3&k:3&k:3&k:3&k:3\\
	     ~&e:1&e:1&e:1&e:1&e:1\\
	     ~&s:2&s:2&s:2&s:2&s:2\\
	     \midrule
	     \multirow{4}{*}{Transformer-5}&c:112&c:112&c:120&c:120&c:128\\
	     ~&d:3&d:3&d:6&d:8&d:8\\  
	     ~&e:1&e:1&e:1&e:1&e:1\\
	     ~&s:2&s:2&s:2&s:2&s:2\\
	     \midrule
	     \multirow{4}{*}{Transformer-6}&c:176&c:176&c:184&c:176&c:184\\
	     ~&d:3&d:3&d:6&d:8&d:8\\  
	     ~&e:1&e:1&e:1&e:1&e:1\\
	     ~&s:1&s:1&s:1&s:1&s:1\\ 
	     \midrule
	     \multirow{5}{*}{Transformer-7}&c:208&c:208&c:216&c:216&c:216\\
	     ~&d:3&d:3&d:6&d:5&d:5\\ 
	     ~&e:1&e:1&e:1&e:1&e:1\\
	     ~&s:2&s:2&s:2&s:2&s:2\\ 
	     \midrule
	     MBPool&1792&1792&1792&1792&1792\\
	     \midrule
	     Resolution&192&224&256&288&288\\
		\bottomrule 
	\end{tabular}}
	\label{nasvit}
\end{table}

\subsection{Visualization Architecture on Dynamic Network}
\begin{figure}[h]
	\centering
	\begin{minipage}{1\linewidth}
		\centerline{\includegraphics[width=\textwidth]{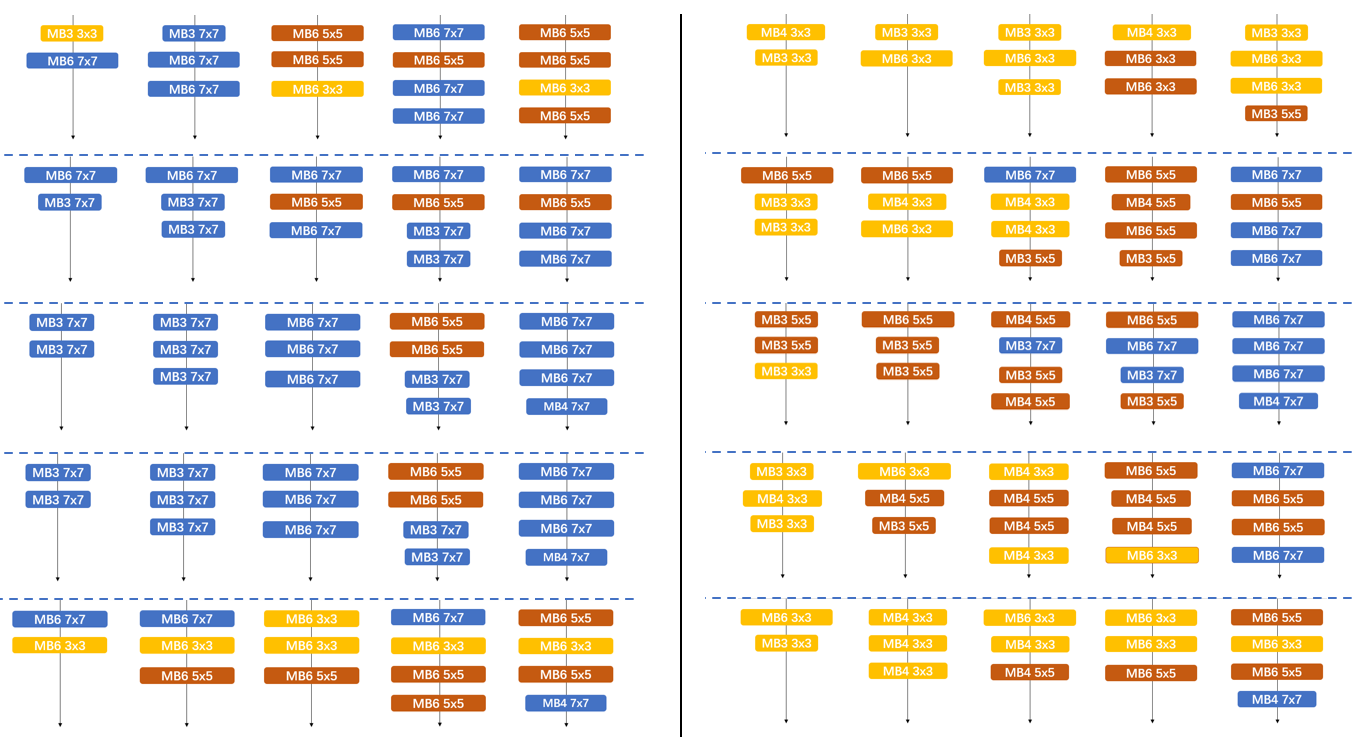}}
	\end{minipage}
	\caption{CPU (right) and GPU (left) Pareto-optimal sub-blocks selected in the MobileNet-V3 super-net. From top to bottom is from block$1$ to block$5$. 
	Blocks in the same row are sub-blocks of the same super-block, and for the same device, the size of the sub-blocks increases sequentially from left to right.}
	\label{afig-blocks}
\end{figure}

Figure \ref{afig-blocks} shows the Pareto-optimal sub-blocks selected on Jetson TX2 CPU/ Raspberry Pi and Jetson TX2 GPU in the MobileNet-v3 search space, respectively. 
GPU prefers shallow and wide block architectures, thus its sub-blocks have fewer layers but larger kernel size and channel number, while CPU prefers deep and narrow block architectures, thus its sub-blocks have more layers, but the kernel size and channel number are smaller.

\section{Discussion}
\subsection{Limitations and Future Work}
In this section, we discuss the limitations of this work.
\begin{itemize}
    \item The theoretical explanation of the law of FLOPs and accuracy changes proposed by MathNAS needs to be strengthened. We intend to continue to explore and try to give a complete theoretical proof from the aspect of network interpretability.
    \item Zero-shot NAS algorithms have been proven to be more efficient. Our future goal is to investigate the potential of applying MathNAS to zero-shot NAS algorithms.
\end{itemize}

\subsection{Potential negative societal impact}
Our proposed technique for rapidly designing network architectures may lead to unemployment of network architecture designers. The technology can also be misused by those who might create evil artificial intelligence.

\end{document}